%% file: main.tex
\documentclass[10pt,twocolumn,letterpaper]{article}

\usepackage{cvpr}              
\usepackage{tablefootnote}
\input{preamble}

\definecolor{cvprblue}{rgb}{0.21,0.49,0.74}
\usepackage[pagebackref,breaklinks,colorlinks,allcolors=cvprblue]{hyperref}
\usepackage{physics}
\usepackage{booktabs}
\usepackage{float}
\usepackage{multicol, multirow}
\usepackage{capt-of,etoolbox}
\usepackage{algorithm2e}
\usepackage{mdframed}

\def\confName{CVPR}
\def\confYear{2025}

\title{Self-Cross Diffusion Guidance for Text-to-Image Synthesis of Similar Subjects}

\author{Weimin Qiu \qquad Jieke Wang \qquad Meng Tang \\
University of California Merced \\ 
{\tt\small \{wqiu5, jwang450, mtang4\}@ucmerced.edu }
}

\makeatletter
\apptocmd\@maketitle{{\teaser \par}}{}{}
\makeatother

\begin{document}

\newcommand{\myteaserwidth}{0.124\textwidth}
\newcommand{\myteasercmd}[2]{\includegraphics[width=\myteaserwidth]{sec/images/#1/sd/#2} \hspace{-4mm} & \includegraphics[width=\myteaserwidth]{sec/images/#1/initno/#2} \hspace{-4mm} & \includegraphics[width=\myteaserwidth]{sec/images/#1/conform/#2} \hspace{-4mm} & \includegraphics[width=\myteaserwidth]{sec/images/#1/selfcross/#2}}

\newcommand{\teaser}{%
\centering
\scalebox{0.95}{
\begin{tabular}{cccccccc}
    \multicolumn{4}{c}{\large \zapchan{a rose and a carnation}} & \multicolumn{4}{c}{\large \zapchan{a leopard and a tiger}} \\
    \hspace{-5mm} \myteasercmd{similar-subjects}{rose_carnation_201084156} & \myteasercmd{similar-subjects}{leopard_tiger_2142450401} \\
    \multicolumn{4}{c}{\large \zapchan{a beagle and a collie}} & \multicolumn{4}{c}{\large \zapchan{a eagle and an owl}} \\
    \hspace{-5mm} \myteasercmd{similar-subjects}{beagle_collie_427291101} & \myteasercmd{similar-subjects}{eagle_owl_457105813} \\
    \scriptsize{SD1.4 \cite{latentdiffusionmodel}} & \scriptsize{INITNO \cite{initno}} & \scriptsize{CONFORM \cite{conform}} & \scriptsize{Self-Cross (Ours)} & \scriptsize{SD1.4 \cite{latentdiffusionmodel}} & \scriptsize{INITNO \cite{initno}} & \scriptsize{CONFORM \cite{conform}} & \scriptsize{Self-Cross (Ours)}
\end{tabular}
}
\vspace{-2mm}
\captionof{figure}{Our Self-Cross guidance addresses subject mixing in particular for similar subjects. Our training-free method can boost the performance of any Unet-based or transformer-based diffusion models such as Stable Diffusion 1, 2, and 3 (shown in Fig.~5 and 6.)}
\label{fig:teaser} %
\vspace{4mm}
}

\maketitle

\input{sec/0_abstract}

\input{sec/1_intro}

\input{sec/2_related}
\input{sec/3_methods}

\input{sec/4_experiments}
\input{sec/conclusion}

\clearpage

\paragraph{Acknowledgement} This research was supported by the Department of Defense under funding award W911NF-24-1-0295 and by Google Cloud research credits program.

{
    \small
    \bibliographystyle{ieeenat_fullname}
    \bibliography{main}
}

\clearpage

\input{sec/X_suppl}

\end{document}

%% file: preamble.tex
\newcommand{\calL}{\mathcal{L}}
\newcommand{\bbE}{\mathbb{E}}

\DeclareMathOperator*{\argmin}{arg\,min}

\newcommand*\zapchan{\fontfamily{pzc}\selectfont}
\usepackage{algorithmic}

%% file: sec/0_abstract.tex
\begin{abstract}
Diffusion models achieved unprecedented fidelity and diversity for synthesizing image, video, 3D assets, etc.
However, subject mixing is an unresolved issue for diffusion-based image synthesis, particularly for synthesizing multiple similar-looking subjects.
We propose Self-Cross Diffusion Guidance to penalize the overlap between cross-attention maps and the aggregated self-attention map.
Compared to previous methods based on self-attention or cross-attention alone, our guidance is more effective in eliminating subject mixing. 
What's more, our guidance addresses subject mixing for all relevant patches beyond the most discriminant one, e.g., the beak of a bird.
For each subject, we aggregate self-attention maps of patches with higher cross-attention values. Thus, the aggregated self-attention map forms a region that the whole subject attends to.
Our training-free method boosts the performance of both Unet-based and Transformer-based diffusion models such as the Stable Diffusion series. 
We also release a similar subjects dataset (SSD), a challenging benchmark, and utilize GPT-4o for automatic and reliable evaluation.
Extensive qualitative and quantitative results demonstrate the effectiveness of our self-cross diffusion guidance.
\end{abstract}

%% file: sec/1_intro.tex
\section{Introduction}
\label{sec:intro}

Diffusion-based generative models have made significant progress in recent years in synthesizing high-quality images~\cite{latentdiffusionmodel}, videos~\cite{sora}, 3D assets~\cite{poole2022dreamfusion}, etc.
With a simple text prompt, diffusion-based generative models such as Stable Diffusion~\cite{latentdiffusionmodel} can create highly photorealistic or artistic images of various styles and subjects. 
Such progress revolutionized many applications including the content creation.
%
%
However, text-to-image diffusion models still have many issues such as \textit{subject neglect}, \textit{subject mixing}, and \textit{attribute binding}.
We are particularly interested in solving \textit{subject mixing} in this work.
Many mitigations~\cite{attendandexcite,initno,conform} were proposed to enhance the faithfulness of text-to-image synthesis but the issue of \textit{subject mixing remains}, see failure cases of previous methods~\cite{initno,conform} in Fig.~\ref{fig:teaser}.
This is more prominent when synthesizing multiple similar-looking subjects, e.g., a photo of a leopard and a tiger in Fig.~\ref{fig:teaser}.

\begin{figure}[!t]
    \centering
    \includegraphics[width=0.8\columnwidth]{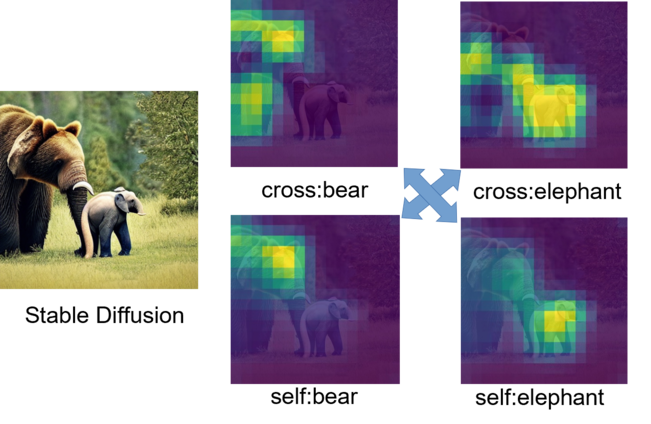} 
    \includegraphics[width=0.8\columnwidth]{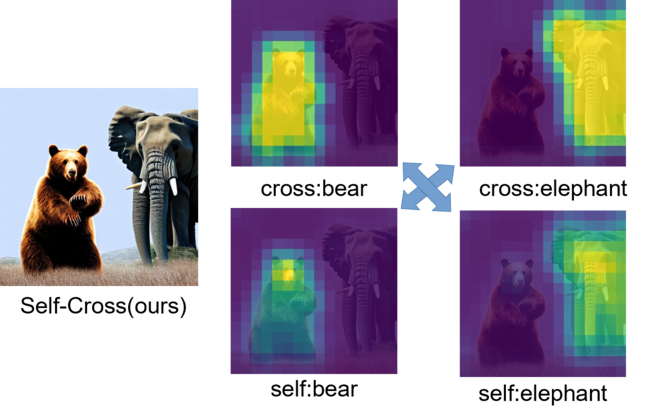} 
    \vspace{-2mm}
    \caption{Results of Stable Diffusion~\cite{latentdiffusionmodel} and our method with Self-Cross guidance for the same prompt "a bear and an elephant". Images are generated from the same random seed. ``cross" means cross-attention map and ``self" means the aggregated self-attention map. The overlap between self-attention and cross-attention leads to subject mixing, while Self-Cross guidance reduces overlapping.}
    \label{fig:motivation}
    \vspace{-3mm}
\end{figure}

Since self-attention maps reflect the similarity of patches and cross-attention maps reflect the subjects.
Intuitively, patches from a subject should not show large similarity to patches from other subjects at least for some timesteps and layers to avoid copying patches from other subjects. 
Empirically, Fig.~\ref{fig:motivation} shows a failure case for Stable Diffusion~\cite{latentdiffusionmodel} with subject mixing and an obvious overlap between cross-attention maps w.r.t. a subject token and self-attention maps w.r.t. patches of another subject.
Our insight is that \textit{a subject should not attend to other subjects by its self-attention maps}.
As shown in Fig.~\ref{fig:motivation}, while overlapping between self-attention and cross-attention maps results in a failure case, our Self-Cross diffusion guidance penalizes such overlaps yielding synthetic images without subject mixing.

Our method is different from previous methods based on cross-attention or self-attention maps alone~\cite{initno,conform, sepen}.
We are the first to explore regularization between self-attention and cross-attention maps.
What's more, we formulate Self-Cross diffusion guidance for self-attention maps of multiple image patches beyond the most discriminant one.
It is well-known that neural networks for perceptron such as image classification tend to focus on the discriminant region of an image, e.g., the beak of a bird. 
{However, apart from the most discriminant patch, other patches with more details are also particularly important for synthesizing multiple similar-looking subjects.
We first adaptively identify all image patches corresponding to a subject and then avoid
their attendance to other subjects by self-attention maps. 
As a training-free method, it outperforms other single-patch-based approaches~\cite {attendandexcite,initno} qualitatively and quantitatively.

Previous datasets with prompts for image synthesis are not challenging in terms of subject mixing. What's worse, the commonly used CLIP score doesn't correlate well with human judgment.
Therefore, to facilitate image synthesis of similar objects, we release a similar-subject dataset (SSD) consisting of text prompts with similar subjects.
To enable benchmarking at scale, we leverage state-of-the-art vision language models such as GPT-4o to evaluate synthetic images of different methods by visual question answering~\cite{tifa}.

The main contributions of this paper are as follows.

\begin{itemize}
  \item We propose Self-Cross diffusion guidance for text-to-image synthesis of similar-looking subjects, which effectively addresses subject mixing as shown in our qualitative and quantitative results.
  \item Our guidance is training-free and can improve the performance of pre-trained models such as Stable Diffusion.
  \item We propose similar-subject dataset (SSD), a new benchmark for image synthesis of similar subjects. Our benchmark includes prompts for two or three similar subjects and an effective metric using VLM~\cite{tifa}.
  \item As a side effect evidenced by improved existence and recognizability scores, our method reduced subject neglect.
\end{itemize}

%% file: sec/2_related.tex
\section{Related Work}
\label{sec:relatedwork}

\paragraph{Text-to-Image Diffusion Models}
Given text prompts, text-to-image synthesis aims to generate visually coherent images.
Early approaches utilized GANs \cite{gan, attngan, dfgan, dmgan, saharia2022photorealistic, stylegant, gigagan} and autoregressive models \cite{ramesh2021zero, rqtransformer, cogview, cogview2, makeascene, muse, cm3leon, yu2022scaling}.
Recently, with ground breaking advancements in diffusion models \cite{Thermodynamics, ddpm, ddim, dhariwal2021diffusion_beat, meng2021sdedit, song2021scorebased_sde, CFG}, the focus of text-to-image synthesis has shifted toward diffusion models \cite{latentdiffusionmodel, saharia2022photorealistic, dalle2, dalle3, cogview3,sdxl,sd3}.
Although diffusion models can generate photorealistic images, ensuring faithful adherence to the provided text prompt remains a significant challenge.
To tackle this issue, methods like ReCap \cite{recap}, DALLE3 \cite{dalle3}, and SD3 \cite{sd3} leverage improved image-caption pairs during training or incorporate multiple language encoders to capture more expressive language representations.
However, these methods require training models from scratch, which entails substantial computational costs and makes them inapplicable to popular models like Stable Diffusion \cite{latentdiffusionmodel} and Imagen \cite{saharia2022photorealistic}. Furthermore, these models only partially address issues such as subject neglect, subject missing, and attribute binding, leaving room for improvement towards prompt-faithful synthesis.


\vspace{-2mm}

\paragraph{Guidance for Consistent Text-to-Image Generations} 
Training-free inference-time optimization is an active research area to improve the consistency of the pre-trained text-to-image models.
These methods typically extract internal representations of the denoising networks and correct the denoising trajectory to improve the alignment to the given prompt.
One way to modulate the denoising trajectory is to replace internal features, such as attention modules in PnP-Diffusion \cite{pnpdiffusion}, FreeControl \cite{mo2024freecontrol}, Prompt-to-Prompt \cite{prompt2prompt}, DenseDiffusion \cite{densediffusion}, and MasaCtrl \cite{masactrl}.
While effective for image editing and style transfer, these methods fail to address critical issues like subject mixing and attribute misalignment.
Another line of work optimizes latents by minimizing guidance loss, in a way similar to classifier guidance \cite{dhariwal2021diffusion_beat} and classifier-gree guidance \cite{CFG}.
%
To address the problems of subject neglect, subject mixing, and attribute binding, many approaches have been proposed.
\citet{richt2i}, Self-Guidance \cite{dsg}, and DisenDiff \cite{disendiff} design loss functions for image editing and controllable generation. Attention Refocusing \cite{attnrefocusing}, BoxDiff \cite{boxdiff}, TokenCompose \cite{tokencompose}, and \citet{richt2i} use pre-defined layouts, either from external models or users, as inference-time supervision. However, these methods rely on prior knowledge, which is sometimes unreachable in real world. 
Therefore, other knowledge-free methods have been developed. 
Following Attend\&Excite \cite{attendandexcite} and A-STAR \cite{astar} steps, CONFORM \cite{conform} takes contrastive loss for subject separation and attribute-binding.
Linguistic Binding \cite{linguisticbinding} introduces a variant of Kullback-Leibler divergence as the loss for improved consistency.
INITNO \cite{initno} leverages both cross-attention maps and self-attention maps to refine the initial noise. 
While they use cross-attention and self-attention separately, we emphasize that the interaction between self-attention and cross-attention is key to eliminating subject mixing and improving faithfulness.
Moreover, these methods typically consider the most discriminant patch, which is insufficient for removing subject mixing.

%% file: sec/3_methods.tex
\section{Preliminaries}


\vspace{-2mm}

\paragraph{Diffusion Model}
Latent diffusion model~\cite{latentdiffusionmodel} operates in a latent space instead of the pixel space, which largely reduces the computational complexity of image generation. An encoder and a decoder are trained to encode images and decode lower-dimensional latents respectively. Furthermore, cross-attention between prompts and image patches allows controllable image generation with various prompts, such as layouts, semantics, and texts.

In the latent space, the forward process gradually adds Gaussian noises on the latent code $z_{0}$ over time until it completely deteriorates to Gaussian noise $z_{T}$. While in the reverse denoising process, a denoising network \cite{unet} $\epsilon_{\theta}$ denoises the latent code $z_{t}$ iteratively until time step zero $z_{0}$. The training objective is formally defined as:
\begin{equation}
\calL = \bbE_{z_{t} ,\epsilon\sim N(0,I),c(y),t} ||\epsilon - \epsilon_{\theta}(z_{t}, c(y), t)||^{2}  
\end{equation}
where $c(y)$ represents the condition embedding, for example CLIP embedding for text prompt $y$.

In cross-attention modules, condition embeddings $c$ are projected to query $Q$ and values $V$. Accordingly, intermediate representations from UNet are projected to keys $K$. Therefore, the cross-attention maps can be described as:
\begin{equation}
A^{c} = \operatorname{Softmax}(\frac{QK^T}{\sqrt{d}}) 
\end{equation}
where $\sqrt d$ is a scaling factor \cite{attn}. Each text token has a cross-attention map with shape $R^{h/P\times w/P}$ for patch size $P\times P$. The cross-attention maps reflect the attendance of text tokens to patches. On the other hand, the self-attention map for each patch indicates the relationship between different patches. We denote the cross attention map of a text token $k$ as $A_k^c\in R^{h/P\times w/P}$ and the self attention map of a patch $(x,y)$ as $A_{x,y}^s\in R^{h/P\times w/P}$.

\paragraph{Attention Based Guidance}

As discussed in Sec.~\ref{sec:relatedwork}, one methodology for consistent text-to-image generation is attenton-based guidance~\cite{attendandexcite,sepen,initno,conform,attnrefocusing} using self attention or cross attention. Here we briefly introduce the most relevant methods to our Self-Cross diffusion guidance.

To avoid subject neglect, Attend-and-Excite~\cite{attendandexcite} finds the patch with maximum cross attention for each token, and penalizes if the maximum cross attention is small. In other words, it encourages the appearance of specified subjects. Speficailly, the cross-attention response score is defined as,

\begin{equation}
S_{\operatorname{cross-attn}} = \max_{k\in K} S_{\operatorname{cross-attn},k}
\end{equation}
where $K$ indicates the set of subjects' tokens and
\begin{equation}
S_{\operatorname{cross-attn},k} = 1 - \max(A_{k}^{c}).
\end{equation}

Unlike Attend-and-Excite\cite{attendandexcite}, our Self-Cross diffusion guidance loss is between self-attention and cross-attention. Besides, it's formulated for all relevant subject patches beyond the most discriminant one.
The most similar guidance to ours is the self-attention conflict score introduced in INITNO~\cite{initno}. However, INITNO~\cite{initno} is still limited to the most discriminant patch, and our method outperformed INITNO~\cite{initno} by a large margin shown in our experiments.
While separation loss~\cite{sepen} penalizes overlap between cross-attention maps, it isn't a training-free method. Our proposed Self-Cross guidance is novel and complementary to existing works on attention-based guidance.

\paragraph{Optimization of Guidance Loss} The standard method for optimizing a guidance loss is through gradient descent. To better minimize a guidance loss, Iterative Latent Refinement \cite{attendandexcite} runs multiple steps of gradient descent until the guidance loss is bellow a threshold. To find a better initial noise, Initial noise optimization~\cite{initno} (INITNO) optimizes over the mean and variance of initial noise util the initial guidance loss is satisfactory.
We adopted the two techniques above for our Self-Cross diffusion guidance.



\section{Our Method}
\label{sec:method}


We describe Self-Cross diffusion guidance given a pair of similar subjects, namely ``a bear and an elephant" with "bear" at index $i=2$ and "elephant" at index at $j=5$.
Our method can be easily extended to multiple pairs.

\begin{figure}[b]
  \centering
  \includegraphics[scale=1.0]{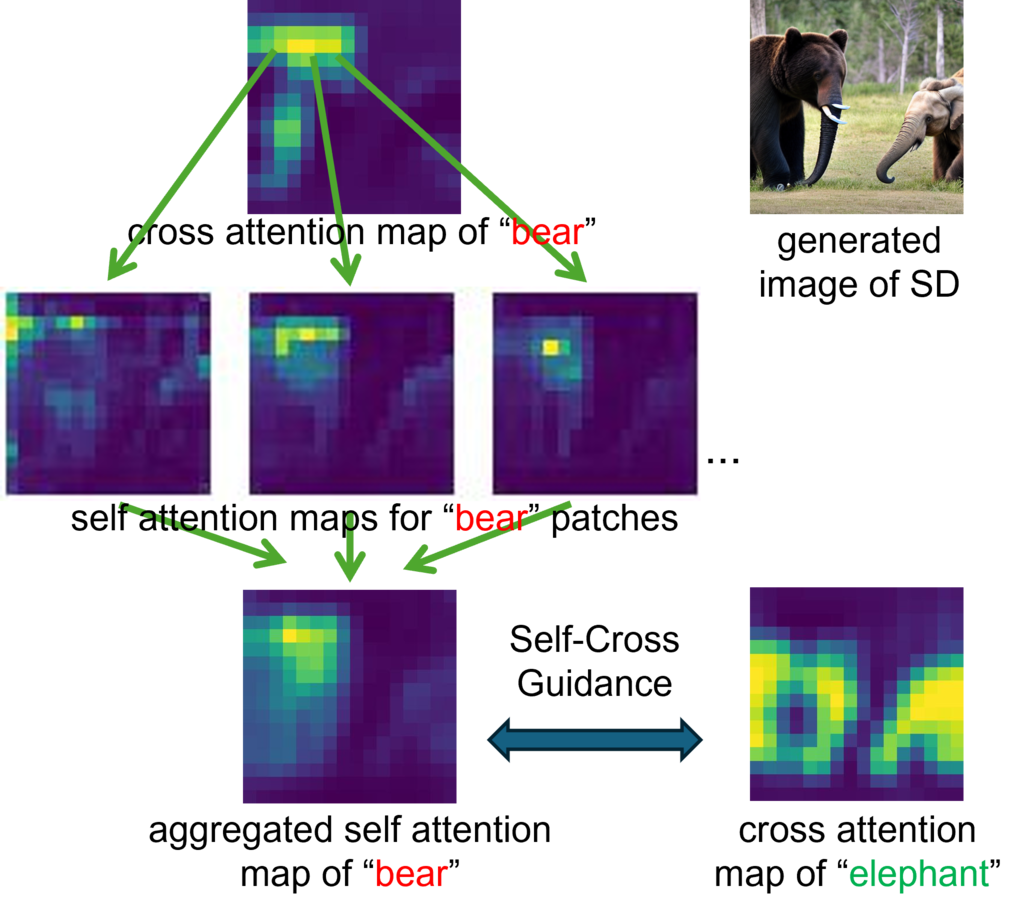}
  \vspace{-4mm}
      \caption{Self-Cross diffusion guidance between the cross-attention of ``elephant" and the self-attention of ``bear"}
\label{fig: Self-Cross guidance}
\label{fig:attention_aggregation}
\vspace{-3mm}
\end{figure}

\paragraph{Aggregation of Self-Attention Maps}
Fig.~\ref{fig:attention_aggregation} describes the aggregation of self-attention maps.
Given the cross-attention map of ``bear", we select patches with high responses and visualize their self-attention maps.
The diversity of self-attention for different patches means the self-attention map for the most discriminant patch alone can't cover all the regions the subject attends to.
Thus, the key to our Self-Cross diffusion guidance is aggregating self-attention maps. 
If limited to the most discriminant patch (with the highest cross-attention value), other foreground patches may lead to subject mixing, as shown in Fig.~\ref{fig:different_patches}.


Firstly, we apply simple and efficient Otsu’s method \cite{otsu} on the cross-attention maps, which automatically returns the threshold and masked patches with relatively higher cross-attention values.
Secondly, we aggregate all the self-attention maps of masked patches to better represent the whole region that a subject attends to.
For subject at $i$ with cross-attention map $A_i^c$, aggregated self-attention map is:


\begin{equation}
A_{i}^{s} = \frac{\sum_{x_{m},y_{n}} (A_{i}^{c}[x_{m},y_{n}]\times A_{x_{m},y_{n}}^{s})}{\sum_{x_{m},y_{n}} A_{i}^{c}[x_{m},y_{n}]}
\end{equation}
Where $[x_{m},y_{n}]$ are the coordinates of the selected patches, $A_{i}^{c}[x_{m},y_{n}]$ is the cross-attention value for token at $i$ of the patch at $[x_{m},y_{n}]$, and $A_{i}^{s}[x_{m},y_{n}]$ is the self-attention map for the patch at $[x_{m},y_{n}]$.

In summary, we select patches corresponding to a subject and then take a weighted sum of the self-attention maps for these patches, where the weights are the patches' cross attention values. Likewise, we obtain aggregated self-attention maps $A_i^s$ for all subjects at given indexes.


\paragraph{Self-Cross Diffusion Guidance}

The aggregated self-attention maps highlight regions that the subject attends to. Our core assumption is that a subject should not attend to other subjects in the image at some time-steps and layers through its self-attention maps. Hence, we propose to penalize the overlap between the aggregated self-attention map of one subject and cross-attention maps of other similar subjects. This has a different effect compared to a loss between cross-attention maps, as aggregated self-attention can be different from cross-attention, shown in Fig.~\ref{fig: Self-Cross guidance}.

The example prompt in Fig.~\ref{fig: Self-Cross guidance} has two pairs of attention maps, aggregated self-attention of ``bear" \& cross-attention of ``elephant" and aggregated self-attention of ``elephant" \& cross-attention of ``bear".
%
The Self-Cross diffusion guidance between the subject at $i$ and the subject at $j$ is defined as their overlap $g(i,j)$,

\begin{equation}
\begin{split}
g(i,j) &= \sum_{x,y} \min(A_{i}^{s}[x,y],A_{j}^{c}[x,y]) \\
&+  \sum_{x,y} \min(A_{i}^{c}[x,y],A_{j}^{s}[x,y])
\end{split}
\end{equation}
where $A_{i}^{s}$ is the self-attention map of subject at $i$ and $A_{j}^{c}$ is the cross-attention map of subject at $j$. 

Now, let's consider more similar subjects.
If there are $N$ similar subjects, then mathematically we would have $C_{N}^{2}$ pairs of subjects. In this case, our Self-Cross diffusion guidance is the average of the $C_{N}^{2}$ ones, as in Eq.\ref{eq:N_pairs},

\begin{equation}
S_{\operatorname{self-cross}} = \sum^{i\neq j}_{i,j\in \Omega} \frac{g(i,j)}{C_{N}^{2}}.
\label{eq:N_pairs}
\end{equation}
where $\Omega$ is the set containing all subject indexes.
Similar to INITNO~\cite{initno}, we also include the cross-attention response score in our total loss.

\begin{equation}
\vspace{-1mm}
    \mathcal{L}_{total}=S_{\operatorname{self-cross}} + \lambda \cdot S_{\operatorname{cross-attn}}.
\label{eq:total_loss}
\vspace{-1mm}
\end{equation}


\paragraph{Spatial Relationship Among Subjects}
We define Our novel self-cross diffusion guidance with much thought to penalize appearance overlap while not hindering generation capabilities, as shown quantitatively in Tab.~\ref{tab:gpt4o-sd2-spatial}, Tab.~\ref{tab:blip} and qualitatively in Fig.~\ref{fig:sd3}. To achieve this,
\begin{enumerate}[label=(\alph*)]
\item We define the guidance only for early timestamps in the reverse process. Because attention maps from early timestamps are known to be semantically meaningful.
\item It is known that attention maps of intermediate layers in diffusion models are highly correlated to semantics. So we enforce our guidance for these selected layers.
\item Other tokens such as verbs and adjectives in a prompt can render subject relationships for image synthesis.
These tokens are not involved in our guidance.
\end{enumerate}
See more implementation details in the appendix A.

\paragraph{Overall Pipeline} Alg.~\ref{alg:selfcross} shows the overall pipeline, which involves initial noise optimization for our total loss~\ref{eq:total_loss}. We apply Self-Cross guidance to the first half of the reverse process, which is more relevant for the semantic structure of synthesized images. Similar to previous work~\cite{attendandexcite,initno}, iterative refinement is conducted to ensure losses are below specified thresholds.

\RestyleAlgo{ruled}

\SetKwComment{Comment}{/* }{ */}

\begin{algorithm}[hbt!]
\caption{T2I with Self-Cross Guidance}\label{alg:selfcross}
\textbf{Input:} A text prompt $P$ and indices $\Omega$ of subjects\\
A pre-trained T2I diffusion model such as $SD(\cdot)$\\
Max iterations: $\tau_{MaxAlterStep}$, $\tau_{MaxIter}$\\
Thresholds: $\tau_{\operatorname{cross-attn}}$, $\tau_{\operatorname{self-cross}}$\\
A set of iterations for refinement $\{t_1,t_2,...,t_k\}$.

\textbf{Output:} Generated image $z^0$.

\vspace{1mm}
\hrule
\vspace{1mm}

\textbf{Noise Initialization Step:} \\
Noise pool $\mathcal{P} \leftarrow \{ \}$, $\operatorname{t} \gets T$\;

\textbf{do INITNO} (loss $\mathcal{L}_{total}=S_{\operatorname{self-cross}} + \lambda S_{\operatorname{cross-attn}}$)\\
\textbf{return} noise pool  $\mathcal{P}$\\
$z^{t} \leftarrow  \argmin_{z^{t} \in \mathcal{P}} \mathcal{L}_{total}$\\
\vspace{1mm}
\hrule
\vspace{1mm}
\textbf{Reverse Process:} \\
\While{$\operatorname{t} \geq \tau_{MaxAlterStep}$}{
    $\triangleright$ Compute attention and losses \\ 
    $S_{\operatorname{cross-attn}}, S_{\operatorname{self-cross}}, {\textunderscore} \leftarrow SD (z^{t}, P, t)$ \\
  \eIf{ $\operatorname{t} \in \{t_1,t_2,...,t_k\}$ and\\ $(S_{\operatorname{cross-attn}} > \tau_{\operatorname{cross-attn}}$ or $S_{\operatorname{self-cross}} > \tau_{\operatorname{self-cross}})$ }
  {
    $\triangleright$ Iterative Refinement Steps \\ 
    $i=0$ \\
    \While{$(S_{\operatorname{cross-attn}} > \tau_{\operatorname{cross-attn}}$ or $S_{\operatorname{self-cross}} > \tau_{\operatorname{self-cross}})$ and $i< \tau_{MaxIter}$}{
        $z^{t} \leftarrow SGD(\mathcal{L}_{total})$
        
        $i=i+1$
    }
  }{
    $z^{t} \leftarrow SGD(\mathcal{L}_{total})$
  }
  $z^{t} \leftarrow SD (z^{t}, P, t)$  \\
  $\operatorname{t}=\operatorname{t}-1$
}
$\triangleright$ Continue the remaining steps without guidance 
\While{$\operatorname{t} > 0$}{
    
    $\textunderscore, \textunderscore, z^{t} \leftarrow SD (z^{t}, P, t)$  \\
    $\operatorname{t}=\operatorname{t}-1$  
}
\textbf{return} $z^0$
\label{alg:selfcross}
\end{algorithm}

%% file: sec/4_experiments.tex
\vspace{-4mm} 
\section{Experiments}

\paragraph{Datasets and Baselines}
Similar to previous work~\cite{attendandexcite,initno, conform} on consistent text-to-image generation, we first report results on three datasets including animal-animal, animal-object, and object-object prompts based on Stable Diffusion models.\footnote{We used the original dataset except for replacing the word ``mouse" with ``rat" for any Animal-Animal prompt including ``mouse". This avoids ambiguity, as ``mouse" can refer to a rodent or a computer mouse.} 
After visualization of experimental results, We found some prompts are visually distinct and not challenging enough.
Hence, we introduce our new dataset \textbf{Similar Subjects Dataset (SSD)} containing 31 prompts with two subjects(SSD-2) and 21 prompts with three subjects(SSD-3). The former ones are designed for primitive baselines, e.g., Stable Diffusion 1, and the latter ones are intended for stronger baselines, e.g., Stable Diffusion 3.
In the new dataset, subjects usually share similar structures with distinguishable details, e.g., different textures for leopards and tigers shown in Fig.~\ref{fig:teaser}.
Qualitative and quantitative results show more subject mixing for each method with the new dataset than the original ones. 
Additionally, We also verify the spatial reasoning capacity of our method on 2D-spatial and 3D-spatial from T2ICompBench~\cite{huang2023t2icompbench}~\cite{huang2025t2icompbench++}. These subsets emphasize spatial relationships between two subjects.

 


We compare our method to the original Stable Diffusion~\cite{latentdiffusionmodel} as well as other training-free methods including Initial Noise Optimization (INITNO)~\cite{initno} and CONFORM~\cite{conform}. Separate-and-Enhance~\cite{sepen} is relevant to our work regarding subject mixing, but it necessitates fine-tuning, making it incomparable to training-free approaches.

\subsection{Qualitative results}

\begin{figure*}
\newcommand{\eachwidth}{0.125\linewidth}
\newcommand{\eachmargin}{\hspace{0mm}}
\newcommand{\pairmargin}{\hspace{-3mm}}
\newcommand{\lefrmargin}{\hspace{-8mm}}
\newcommand{\tltmargin}{\hspace{-4mm}}
\newcommand{\myimgrowAA}[2]{\includegraphics[width=\eachwidth]{sec/images/animal-animal/sd/#1} & \pairmargin \includegraphics[width=\eachwidth]{sec/images/animal-animal/sd/#2} & \includegraphics[width=\eachwidth]{sec/images/animal-animal/initno/#1} & \pairmargin \includegraphics[width=\eachwidth]{sec/images/animal-animal/initno/#2} & \includegraphics[width=\eachwidth]{sec/images/animal-animal/conform/#1} & \pairmargin \includegraphics[width=\eachwidth]{sec/images/animal-animal/conform/#2} & \includegraphics[width=\eachwidth]{sec/images/animal-animal/selfcross/#1} & \pairmargin \includegraphics[width=\eachwidth]{sec/images/animal-animal/selfcross/#2}}
\newcommand{\myimgrowSS}[2]{\includegraphics[width=\eachwidth]{sec/images/similar-subjects/sd/#1} & \pairmargin \includegraphics[width=\eachwidth]{sec/images/similar-subjects/sd/#2} & \includegraphics[width=\eachwidth]{sec/images/similar-subjects/initno/#1} & \pairmargin \includegraphics[width=\eachwidth]{sec/images/similar-subjects/initno/#2} & \includegraphics[width=\eachwidth]{sec/images/similar-subjects/conform/#1} & \pairmargin \includegraphics[width=\eachwidth]{sec/images/similar-subjects/conform/#2} & \includegraphics[width=\eachwidth]{sec/images/similar-subjects/selfcross/#1} & \pairmargin \includegraphics[width=\eachwidth]{sec/images/similar-subjects/selfcross/#2}}
\centering
\scalebox{0.8}{
\begin{tabular}{ccc|cc|cc|cc}
& \multicolumn{2}{c|}{SD1.4 \cite{latentdiffusionmodel}} & \multicolumn{2}{c|}{INITNO \cite{initno}} & \multicolumn{2}{c|}{CONFORM \cite{conform}} & \multicolumn{2}{c}{Self-Cross (Ours)}  \\ \midrule \midrule
%
%
\lefrmargin \multirow{2}{*}{\rotatebox[origin=c]{90}{\makebox[0pt][c]{\hspace{-1mm}\vspace{3mm}\zapchan{a bird and a rabbit}}}} & \tltmargin \myimgrowAA{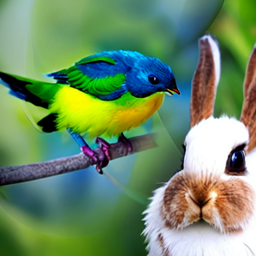}{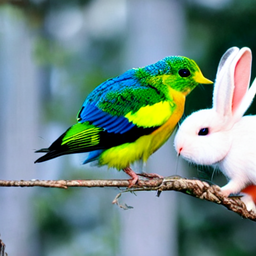} \\
& \tltmargin \myimgrowAA{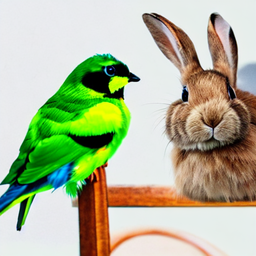}{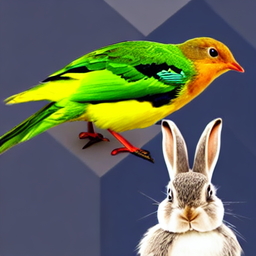} \\ \midrule \midrule
\lefrmargin \multirow{2}{*}{\rotatebox[origin=c]{90}{\makebox[0pt][c]{\hspace{-1mm}\vspace{5mm}\zapchan{a hummingbird and a kingfisher}}}} & \tltmargin \myimgrowSS{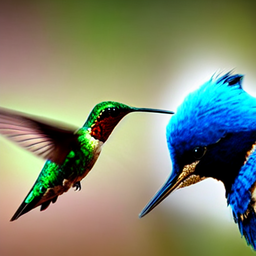}{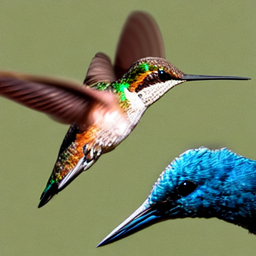} \\
& \tltmargin \myimgrowSS{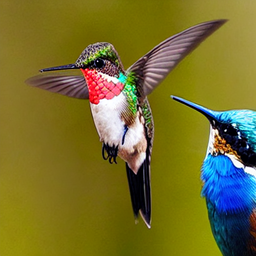}{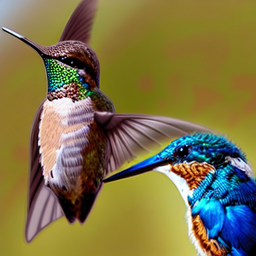} \\ \midrule \midrule
%
%
\end{tabular}}
\vspace{-3mm}
\caption{Qualitative comparisons of Self-Cross (ours) to SD1.4 \cite{latentdiffusionmodel}, INITNO \cite{initno}, CONFORM \cite{conform}. For each prompt in the left column, we sample four seeds and show the results of different methods.}
\label{fig:qualitative-comparisons}
\vspace{-2mm}
\end{figure*}

\begin{figure}
\newcommand{\eachwidth}{0.25\linewidth}
\newcommand{\eachmargin}{\hspace{0mm}}
\newcommand{\pairmargin}{\hspace{-3mm}}
\newcommand{\lefrmargin}{\hspace{-5mm}}
\newcommand{\tltmargin}{\hspace{-2mm}}
\newcommand{\imgrow}[2]{\includegraphics[width=\eachwidth]{sec/images/SD2/original/#1} \includegraphics[width=\eachwidth]{sec/images/SD2/original/#2} &  \includegraphics[width=\eachwidth]{sec/images/SD2/ours/#1} \includegraphics[width=\eachwidth]{sec/images/SD2/ours/#2}}
\scalebox{0.8}{ \lefrmargin
\begin{tabular}{cc|c}
     & SD2.1 & Self-Cross (Ours) \\ \midrule \midrule
    \multirow{2}{*}{\rotatebox[origin=c]{90}{\makebox[0pt][c]{\hspace{-1mm}\vspace{5mm}\zapchan{a bear and a cat}}}} 
    & \tltmargin \imgrow{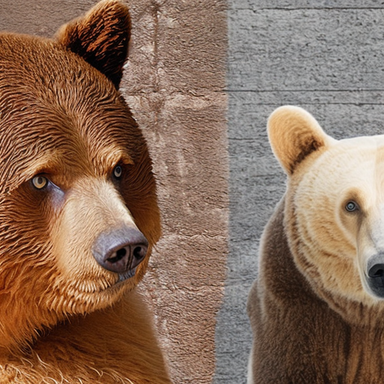}{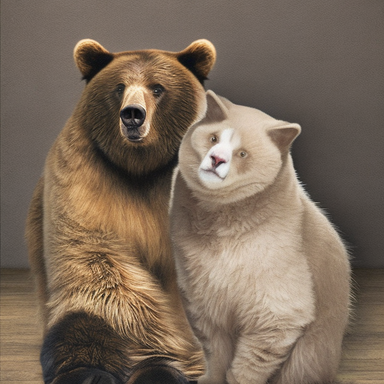}  \\
    & \tltmargin \imgrow{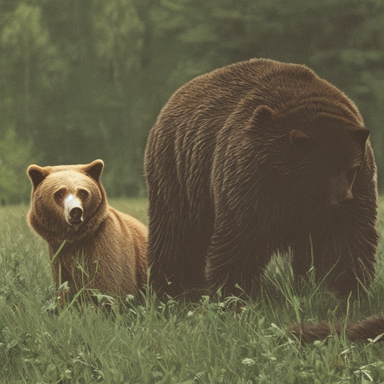}{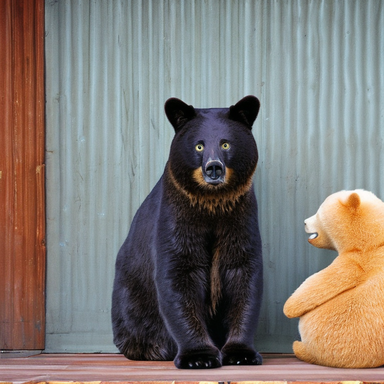} \\ \midrule
    \multirow{2}{*}{\rotatebox[origin=c]{90}{\makebox[0pt][c]{\hspace{-1mm}\vspace{5mm}\zapchan{a bird and a rat}}}}
    & \tltmargin \imgrow{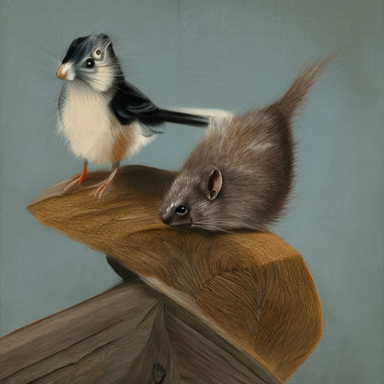}{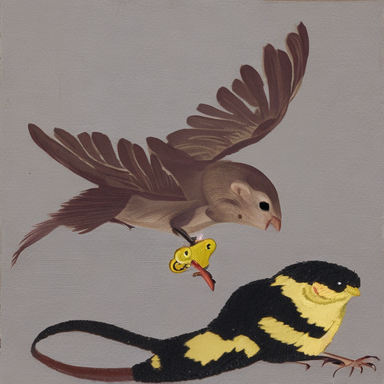}  \\
    & \tltmargin \imgrow{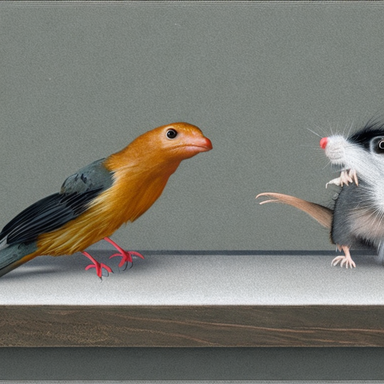}{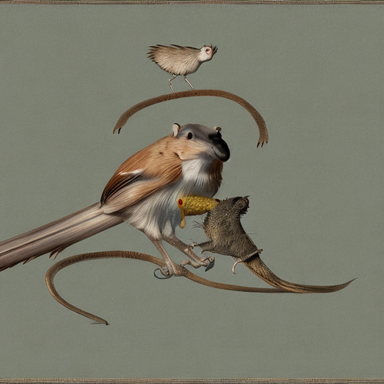} \\ \midrule
    \multirow{2}{*}{\rotatebox[origin=c]{90}{\makebox[0pt][c]{\hspace{-1mm}\vspace{5mm}\zapchan{a dog and a lion}}}}
    & \tltmargin \imgrow{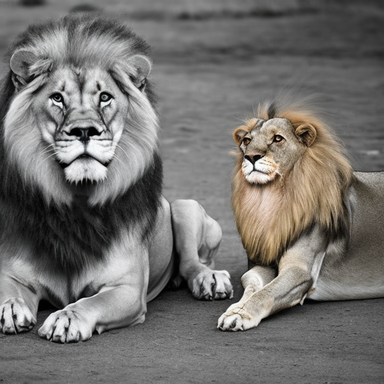}{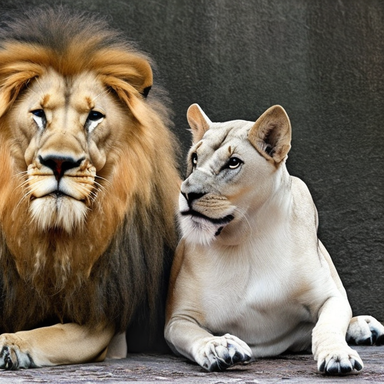}  \\
    & \tltmargin \imgrow{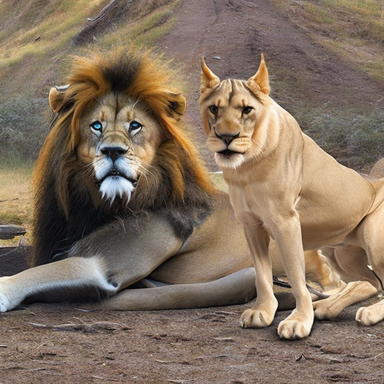}{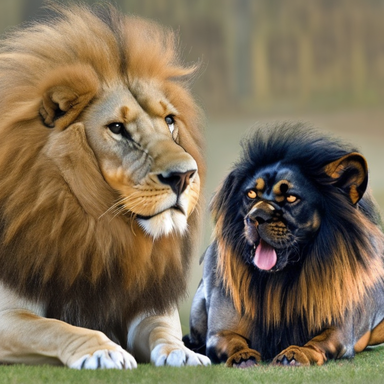} \\
\end{tabular}
}
\caption{Quantitative comparisons between original SD2.1 \cite{latentdiffusionmodel} and our method. 
}
\label{fig:sd2}
\end{figure}

\begin{figure}
\newcommand{\eachwidth}{0.25\linewidth}
\newcommand{\eachmargin}{\hspace{0mm}}
\newcommand{\pairmargin}{\hspace{-3mm}}
\newcommand{\lefrmargin}{\hspace{-5mm}}
\newcommand{\tltmargin}{\hspace{-2mm}}
\newcommand{\imgrow}[2]{\includegraphics[width=\eachwidth]{sec/SD3/original/#1} \includegraphics[width=\eachwidth]{sec/SD3/original/#2} &  \includegraphics[width=\eachwidth]{sec/SD3/ours/#1} \includegraphics[width=\eachwidth]{sec/SD3/ours/#2}}
\scalebox{0.8}{ \lefrmargin
\begin{tabular}{cc|c}
     & SD3-medium & Self-Cross (Ours) \\ \midrule \midrule
    \multirow{2}{*}{\rotatebox[origin=c]{90}{\makebox[0pt][c]{\hspace{-17mm}\vspace{5mm}\zapchan{a bear riding an elephant with a rabbit}}}} 
    & \tltmargin \imgrow{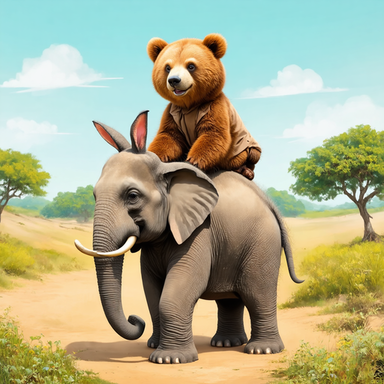}{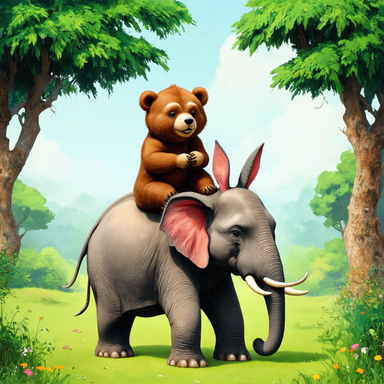}  \\
    & \tltmargin \imgrow{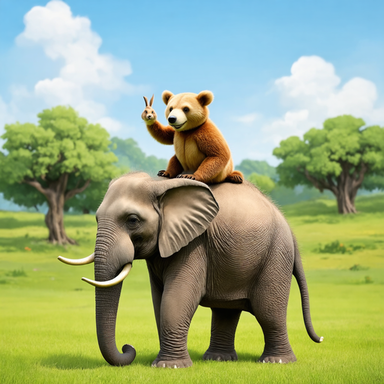}{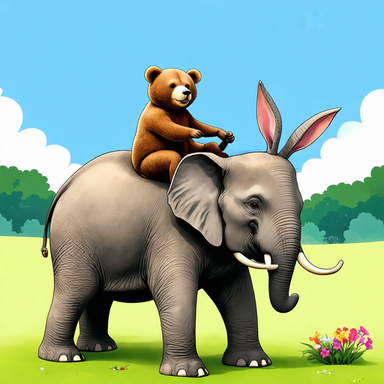}  \\
    & \tltmargin \imgrow{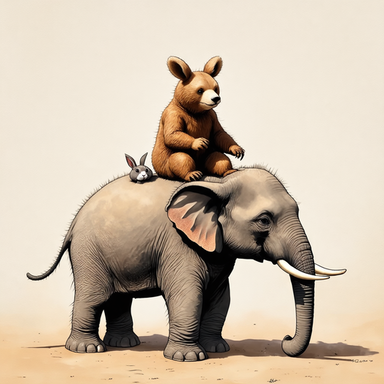}{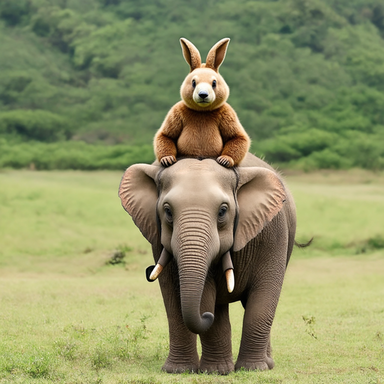} \\ \midrule
    \multirow{2}{*}{\rotatebox[origin=c]{90}{\makebox[0pt][c]{\hspace{-10mm}\vspace{5mm}\zapchan{a beagle and a collie and a husky}}}}
    & \tltmargin \imgrow{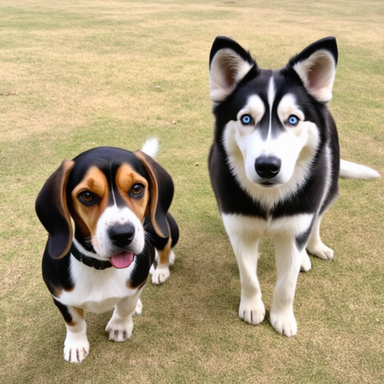}{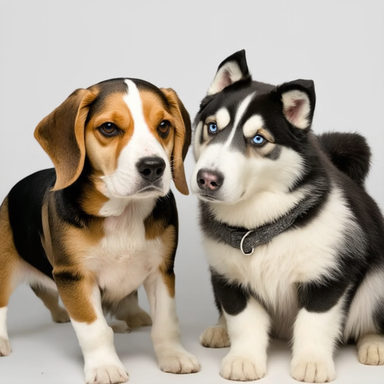}  \\
    & \tltmargin \imgrow{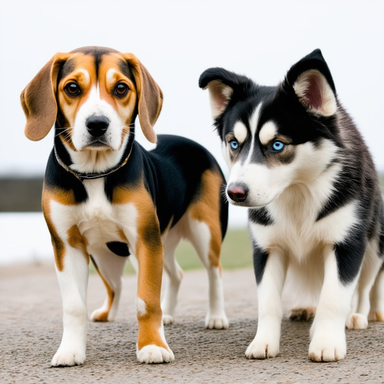}{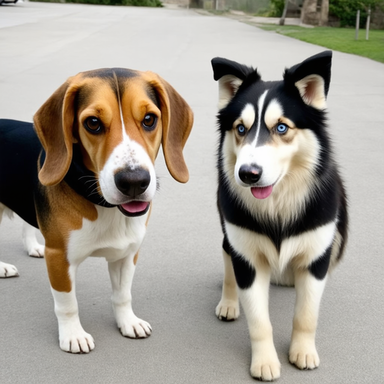}  \\
    & \tltmargin \imgrow{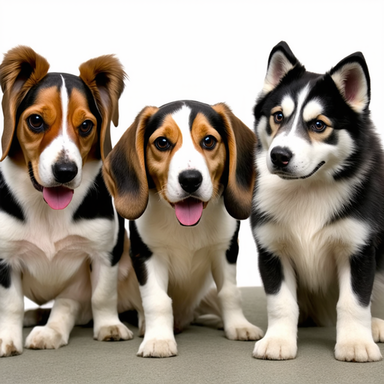}{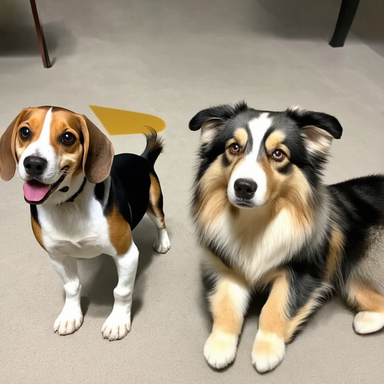} \\ \midrule
\end{tabular}
}
\caption{Quantitative comparisons between SD3-medium \cite{latentdiffusionmodel} and our method. 
}
\label{fig:sd3}
\end{figure}


\paragraph{Self-Cross \textit{v.s.} baselines} We provide the qualitative comparisons in \autoref{fig:qualitative-comparisons}, \autoref{fig:sd2}, and \autoref{fig:sd3} using SD1.4, SD2.1, and SD3-medium respectively.
For each prompt, we used the same list of random seeds for all methods.
Self-Cross diffusion guidance successfully addressed the issue of subject mixing in most cases.
For example, in \autoref{fig:qualitative-comparisons}, given the prompt ``a bird and a rabbit", Stable diffusion~\cite{latentdiffusionmodel}, INITNO~\cite{initno}, and CONFORM~\cite{conform} generated birds with rabbits'ears, while our method generated faithful images without subject mixing. 
Our method synthesized better images for extremely similar subjects too, e.g., ``a hummingbird and a kingfisher". 
Hummingbirds are recognized for their iridescent plumage and long, slender beaks, whereas kingfishers typically feature bold, vibrant colors like blue and have shorter, stout beaks. Appendices E and G show more qualitative results, failure cases, and discussion.



\paragraph{Self-attention and Cross-attention maps} Fig.~\ref{fig:attention_aggregation} presents a cross-attention map and multiple self-attention maps for top-responsive image patches, along with an aggregated self-attention map highlighting the region the entire subject(``bear") attend to. Note that the aggregated self-attention map is different from the original cross-attention map, which motivated Self-Cross diffusion guidance rather than using cross-attention maps alone. The next section explores different self-attention aggregation schemes and their impact on image synthesis.

\paragraph{Self-Cross guidance with the increasing number of patches} To see the effect of self-attention aggregation, we compare our Self-Cross diffusion guidance using different patches. Specifically, given a cross-attention map, we select

\begin{enumerate}[label=(\alph*)]
\item the patch with the maximum cross-attention value.
\item the top 16 patches w.r.t. cross-attention value.
\item masked patches which typically have more than 16 patches. The details of masking are in Sec.~\ref{sec:method}.
\end{enumerate}

\begin{figure}[t]
\centering
\footnotesize
\scalebox{0.9}{
\begin{subfigure}{.13\textwidth}
  \centering
  \includegraphics[scale=0.24]{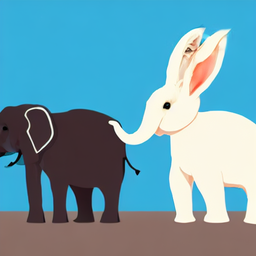}
  \includegraphics[scale=0.24]{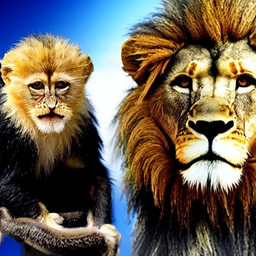}
  \includegraphics[scale=0.24]{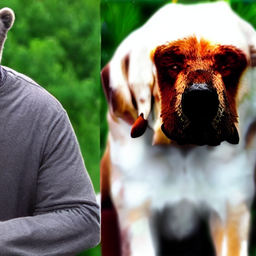}
  \caption{1 patch}
  \label{fig:sfig1}
\end{subfigure}%
\begin{subfigure}{.13\textwidth}
  \centering
  \includegraphics[scale=0.24]{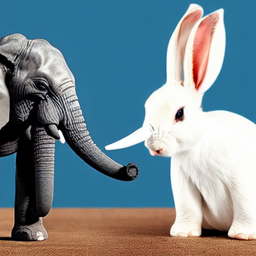}
  \includegraphics[scale=0.24]{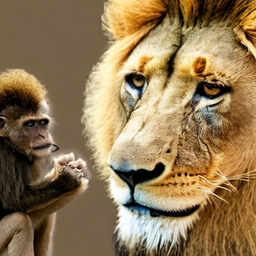}
  \includegraphics[scale=0.24]{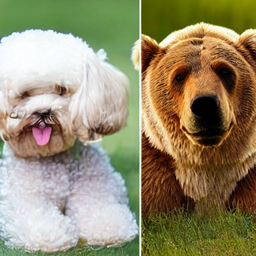}
  \caption{16 patches}
  \label{fig:sfig1}
\end{subfigure}%
\begin{subfigure}{.13\textwidth}
  \centering
  \includegraphics[scale=0.24]{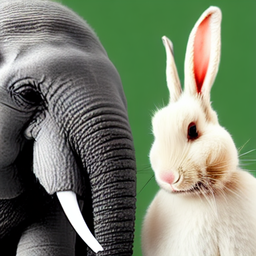}
  \includegraphics[scale=0.24]{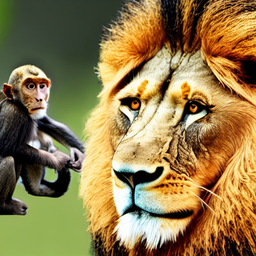}
  \includegraphics[scale=0.24]{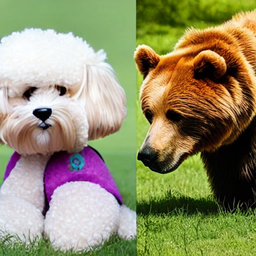}
  \caption{masked patches}
  \label{fig:sfig1}
\end{subfigure}
}
\vspace{-2mm}
\caption{Our Self-Cross guidance works the best with masked patches, which verifies our assumption that all patches of a subject not just the most discriminant one need to be considered for eliminating subject mixing.}
\label{fig:different_patches}
\vspace{-2mm}
\end{figure}

Fig.~\ref{fig:different_patches} shows results with three example prompts, which clearly demonstrate less subject missing with more patches. If only the patch with the maximum cross-attention is used in Self-Cross guidance, the subjects are still mixing. The mixing is less severe but not eliminated with 16 patches, for example mixing in terms of hair, teeth, and feet. The best is when we use all masked patches in the otsu mask and define an aggregated self-attention map for our guidance, in which case the sutle mixing is eliminated.

\begin{table}[h]
    \centering
    \footnotesize
    \begin{tabular}{|c|c|c|c|c|c|}
        \toprule
        Methods & \multicolumn{4}{c|}{SSD-3} \\
        \midrule
        Question Types & Ext & Rec & w/o M & t-t sim \\ \hline %
        
        SD3 Medium  
        & 33.54 & 30.31 & 70.08 & 73.82 \\ \hline

        Self-Cross(Ours) 
        & \textbf{57.92} & \textbf{53.08} & \textbf{77.15} & \textbf{74.96} \\ \bottomrule
    \end{tabular}
    \vspace{-3mm}
    \caption{Quantitative results using SD3 for prompts with three subjects.}
    \label{tab:sd3}
    \vspace{-4mm}
\end{table}

    \begin{table}[h]
        \centering
        \footnotesize
        \scalebox{0.9}{
        \begin{tabular}{|c|c|c|c|c|}
        \toprule
            \% & Animal-Animal & Animal-Obj & Obj-Obj & SSD-2 \\ \midrule 
            SD1.4 & 76.5 & 79.2 & 76.4 & 70.0 \\
            INITNO & 82.2 & 84.0 & 82.3 & 72.0 \\
            CONFORM & 81.9 & 84.6 & 82.0 & 70.7 \\
            Self-Cross(Ours)& \textbf{84.3} & \textbf{84.7} & \textbf{82.5} & \textbf{73.6} \\ \bottomrule
        \end{tabular}}
        \vspace{-2mm}
        \caption{Average text-text similarities ($\uparrow$) on different methods}
        \label{tab:blip}
        \vspace{-3mm}
    \end{table}

    \begin{table}[h]
        \centering
        \footnotesize
        \scalebox{0.85}{
        \begin{tabular}{|c|c|c|c|c|}
        \toprule
            \% & Animal-Animal & SSD-2 & 2D-Spatial & 3D-Spatial \\ \midrule 
            SD2.1 & 81.68 & 73.01 &78.71 & 77.73\\
            INITNO & 83.15 & 73.20 & - & - \\
            Self-Cross(Ours) & 84.02 & 73.53 & \textbf{80.4} & \textbf{80.0}\\ 
            CONFORM & \underline{85.21} & \underline{73.87} & - & - \\
            CONFORM+Ours & \textbf{85.88} & \textbf{74.91} & - & - \\ \bottomrule
        \end{tabular}}
        \vspace{-2mm}
        \caption{Average text-text similarities ($\uparrow$) on different methods}
        \label{tab:blip2}
        \vspace{-3mm}
    \end{table}

    \begin{table*}[tbhp]
        \centering
        \footnotesize
        \scalebox{0.85}{
        \begin{tabular}{|c|c|c|c|c|c|c|c|c|c|c|c|c|}
        \toprule
        Methods & \multicolumn{3}{c|}{Animal-Animal} & \multicolumn{3}{c|}{Animal-Object} & \multicolumn{3}{c|}{Object-Object} & \multicolumn{3}{c|}{SSD-2} \\
        \midrule
        Question Types & Ext & Rec & w/o M %
        & Ext & Rec & w/o M %
        & Ext & Rec & w/o M %
        & Ext & Rec & w/o M \\ \hline %
        
        SD1.4 \cite{latentdiffusionmodel} 
        & 39.51 & 29.70 & 72.24
        & 67.84 & 53.72 & 90.36 %
        & 34.15 & 31.89 & 94.22 %
        & 30.77 & 28.09 & 77.47 \\ \hline

        INITNO \cite{initno} 
        & 89.39 & 77.09 & 82.26
        & 98.37 & \textbf{78.00} & 95.97 %
        & \underline{96.20} & \underline{90.42} & 95.17 %
        & 61.34 & 55.53 & \underline{79.70} \\ \hline 

        CONFORM \cite{conform} 
        & \underline{89.63} & \underline{78.00} & \underline{84.22}
        & 98.37 & 75.09 & \underline{97.16} %
        & 78.58 & 73.12 & \underline{97.48} %
        & \underline{67.54} & \underline{59.90} & 79.50 \\ \hline

        Self-Cross(Ours) 
        & \textbf{94.55} & \textbf{87.79} & \textbf{92.94} 
        & \textbf{99.60} & \underline{75.17} & \textbf{98.30} 
        & \textbf{98.95} & \textbf{93.19} & \textbf{98.65} 
        & \textbf{77.67} & \textbf{70.92} & \textbf{86.45} \\ 
        \bottomrule
        \end{tabular}}
        \vspace{-2mm}
        \caption{TIFA-GPT4o Scores ($\uparrow$) on four benchmarks: Animal-Animal, Animal-Object, Object-Object, and our proposed Similar Subjects Dataset.
        Inspired by TIFA \cite{tifa}, we employ GPT4o \cite{gpt4} as the VQA model to evaluate three aspects: Existence of both subjects (Ext), Recognizability of both subjects (Rec), and Absence of Mixing of subjects (w/o M).
        GPT4o is prompted with several True/False questions, and we report the percentage of True responses as the scores. The list of question prompts is given in the supplementary materials.
        Best results are highlighted in \textbf{bold} and second best results are shown with \underline{underline}.
        }
        \label{tab:gpt4o}
        \vspace{-2mm}
    \end{table*}

    \begin{table*}
        \centering
        \footnotesize
        \scalebox{0.85}{
        \begin{tabular}{|c|c|c|c|c|c|c|c|c|c|c|c|c|c|c|}
        \toprule
        Methods & \multicolumn{3}{c|}{Animal-Animal} & \multicolumn{3}{c|}{SSD-2} & \multicolumn{4}{c|}{2D-Spatial} & \multicolumn{4}{c|}{3D-Spatial} \\
        \midrule
        Question Types & Ext & Rec & w/o M %
        & Ext & Rec & w/o M 
        & Ext & Rec & Rel & w/o M 
        & Ext & Rec & Rel & w/o M \\ \hline %
        
        SD2.1 \cite{latentdiffusionmodel} 
        & 61.63 & 38.23 & 67.51 %
        & 52.84 & 36.98 & 84.93\tablefootnote{vanilla SD2.1 usually generates only one subject of the prompt so its `w/o M' score is high.} 
        & 76.16 & 73.37 & 32.46 & 90.21 
        & 69.1 & 66.17 & 47.72 & 90.33 \\ \hline

        INITNO \cite{initno} 
        & 80.96 & 47.18 & 63.80 %
        & 68.70 & 46.10 & 73.85
        & - & - & - & - 
        & - & - & - & -  \\ \hline 

        Self-Cross(Ours) 
        & 89.53 & 55.03 & 78.79 %
        & 77.35 & 45.36 & 77.08
        & \textbf{87.37} & \textbf{83.90} & \textbf{35.48} & \textbf{91.24}
        & \textbf{87.07} & \textbf{82.60} & \textbf{57.96} & \textbf{91.27} \\ \hline 

        
        CONFORM \cite{conform} 
        & \underline{96.88} & \textbf{70.19} & \underline{92.49}%
        & \textbf{83.71} & 53.90 & 89.76
        & - & - & - & - 
        & - & - & - & - \\ \hline 

        CONFORM+Ours
        & \textbf{97.83} & \underline{69.00} & \textbf{94.66} 
        & 81.39 & \textbf{55.98} & \textbf{93.20} 
        & - & - & - & - 
        & - & - & - & - \\
        \bottomrule
        \end{tabular}}
        \caption{Quantitative benchmarks with SD2.1
        TIFA-GPT4o Scores ($\uparrow$) on two challenging benchmarks: Animal-Animal and our proposed Similar Subjects Dataset.
        Best results are highlighted in \textbf{bold} and second best results are shown with \underline{underline}.
        vanilla SD2.1 usually generates only one subject of the prompt so its `w/o M' score is high.}
        \label{tab:gpt4o-sd2-spatial}
    \end{table*}

\subsection{Quantitative results}
Quantitative results include TIFA-GPT4o scores and text-text similarities.
Our TIFA-GPT4o is a variant of the recently proposed TIFA metric~\cite{tifa}, which is much more correlated with human judgment compared to the widely used CLIP score.
Our main quantitative results are TIFA-GPT4o scores, while CLIP scores should be interpreted with caution (see appendix D for details).

\vspace{-2mm}
\paragraph{TIFA-GPT4o Scores} TIFA \cite{tifa} aims to assess the faithfulness of the generated image to the input prompt. It uses a vision-language model to perform Visual Question Answering (VQA) on the generated image, guided by specific questions about the image's contents. In our implementation, we leverage a more advanced vision-language model, GPT4o \cite{gpt4}, to provide more precise feedback. Given a text prompt in the form of ``a class A and a class B'', we developed questions to (1) verify the existence of both subjects; (2) confirm the recognizability of both subjects, ensuring no artifacts or distortions are present; and (3) check that the image does not exhibit a mixture of the two subjects, i.e., subject mixing.
These three aspects capture both local and global information about the image, serving as comprehensive and reliable metrics including \textbf{Ext}, \textbf{Rec}, \textbf{w/o M}, and \textbf{Rel} scores respectively. W/o score is most relevant to our study of subject mixing, while Ext and Rec scores are effective in assessing the issue of subject neglect. Rel measures the spatial relationship of subjects. The full set of question prompts we designed are in the appendix C.

As shown in \autoref{tab:gpt4o}, when built on Stable Diffusion 1.4, our method is overwhelmingly better than baselines including SD~\cite{latentdiffusionmodel}, INITNO~\cite{initno}, and CONFORM~\cite{conform} on Animal-Animal and Similar Subjects datasets(SSD-1), especially for reducing subject mixing (w/o M score). For animal-animal prompts, we achieved a w/o M score of \textbf{92.94\%}, which is \textbf{8.7\%} better than the second-best score (84.22\% CONFORM). For similar subjects dataset(SSD-1), our w/o M score of \textbf{86.45\%} is \textbf{6.7\%} better than the second-best score (70.70\% INITNO).
Besides, for less challenging datasets with distinct subjects including Animal-Object and Object-Object, we still achieved better results.

Interestingly, although our method focuses on addressing $subject mixing$, it can also reduce $subject neglect$ and improve the recognizability/fidelity of generated subjects to some extent.
However, visualization of the attention maps shows the inefficiency of $attend\&excite$ in Stable Diffusion 2.1. As we rely on $attend\&excite$ to encourage the existence of different subjects, our performance is influenced in Stable Diffusion 2.1.
So, for Stable Diffusion 2.1, we apply CONFORM for the first few steps to ensure the existence of subjects and apply our method in later steps(indicated as CONFORM+ours), as shown in Tab.~\ref{tab:gpt4o-sd2-spatial}. To further prove the effectiveness of our method on DiT, we also implement our method on Stable Diffusion 3 medium as shown quantitatively in Tab.~\ref{tab:sd3} and qualitatively in Fig.~\ref{fig:sd3}.

\paragraph{Text-text similarities}
Text-text similarities, also known as BLIP scores, are the similarity between captions generated by a vision-language foundation model \cite{blip} and the original prompts used to synthesize the images. This metric captures subjects and attributes from the original prompt, then measures the coherency and consistency of the generated content with the textual descriptions. As shown in \autoref{tab:blip}, \autoref{tab:blip2}, and \autoref{tab:sd3}, compared with other methods, our approach, including the combined one, reaches SOTA performances on all benchmarks.



%% file: sec/conclusion.tex
\section{Conclusion and Future Work}

Subject mixing remains a persistent issue for diffusion-based image synthesis, particularly for similar-looking subjects.
We propose Self-Cross diffusion guidance to boost the performance of any diffusion-based image synthesis of similar subjects.
Our method is motivated by the overlap between self-attention maps and cross-attention maps for mixed subjects, which is penalized by the proposed self-cross diffusion guidance loss during inference.
Further more, we aggregate self-attention maps for multiple patches to a single attention map.
In other words, our formulation involves all relevant patches of a subject beyond the most discriminant one.
We are the first to reduce overlap between cross-attention and aggregated elf-attention maps, while previous methods are limited to the self-attention map from one patch or rely on cross-attention maps alone for guidance.
We utilize standard gradient-based optimization and initial noise optmization~\cite{initno} for minimizing our guidance loss during inference.
We also released SSD, a new dataset of similar subjects for image synthesis, and leveraged the latest vision large language model (GPT-4o) for automatic and reliable evaluation of different methods.
Qualitative and quantitative results show significant improvement over previous approaches.
Our Self-Cross guidance greatly reduced subject mixing while also reducing the issue of object neglect as a side effect.

In the future, we will extend our approach to video generation of similar subjects that face challenges of subject mixing. We also anticipate the issue of subject mixing to be less prominent with newer backbone models but not disappear. We will keep exploring variants of Sef-Cross diffusion guidance for finer-grained subjects synthesis and addressing other issues such as attribute binding.

%% file: sec/X_suppl.tex
\clearpage
\setcounter{page}{1}
\maketitlesupplementary

\appendix

\section{More implementation details}

\begin{figure}
\centering
    \includegraphics[scale=0.36]{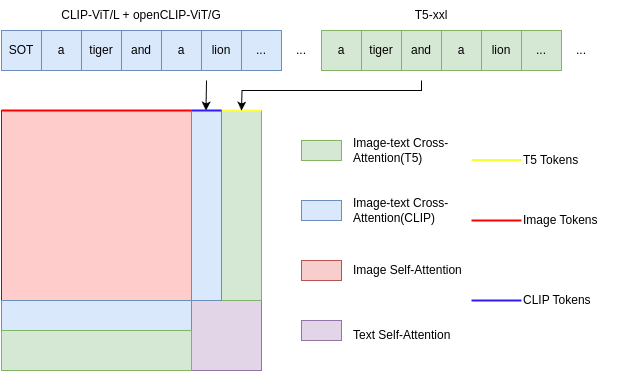} 
    \vspace{-1.5mm} \\
    \caption{\footnotesize{attention maps in multimodal diffusion transformer.}}
    \label{fig:sd3attn}
    \vspace{-6mm}
\end{figure}

Our method is training-free. Following the setting from Attend\&Excite \cite{attendandexcite}, we use pseudo-numerical methods \cite{plmss} and classifier-free guidance \cite{CFG} to generate images with the original image resolutions of Stable Diffusion models. We apply Self-Cross Diffusion Guidance to the first half(25 steps) of the sampling process(50 steps in total). Empirically, we apply refinements at the 10th and 20th steps of the sampling process with thresholds of 0.2 for the cross-attention response score $S_{\operatorname{cross-attn}}$ and 0.3 for self-cross guidance $S_{\operatorname{self-cross}}$. For each prompt, we generated 65 images with consistent random seeds for each method. Tab.~\ref{tab:dataset} shows the number of prompts and generated images for our experiments.  

UNet-based Diffusion models have attention maps of different resolutions including ${16\times 16, 24\times 24, 32\times 32}$, etc. We chose the attention maps that were found most semantically meaningful. In Stable Diffusion 1, we chose attention maps with a resolution of 16 × 16 \cite{ptp}. In Stable Diffusion 2, We empirically chose attention maps with a resolution of 24 × 24. Note that for cross-attention maps of sizes larger than 16 × 16, we normalize their sum to 1 so the values in self-attention maps won't be too small in comparison.

For diffusion models based on multimodal diffusion transformers, e.g. Stable Diffusion 3-medium, we replace the conventional cross-attention with the part of attention between text tokens and image tokens(image-text attention) and replace self-attention with the attention between image tokens(image self-attention) \cite{sd3} \cite{EnhancingMMDiT}. All of these can be extracted from the multimodal diffusion transformer module, as shown in ~\ref{fig:sd3attn}.  As SD3-medium concatenates the text embeddings from CLIP and T5, we extract the corresponding two image-text attention maps and take the maximum value of the two for each image token (patch) to build the cross-attention map for Self-Cross Diffusion Guidance. Eq.\ref{eq:sd3attn}.

\begin{equation} 
A_{i}^{c}[x,y] = \max(A_{i,clip}^{c}[x,y],A_{i,t5}^{c}[x,y]) 
\label{eq:sd3attn}
\end{equation}

After attention maps were extracted, We averaged attention maps head-wise and layer-wise in our implementation.

    \begin{table}[h]
        \centering
        \footnotesize
        \scalebox{0.9}{
        \begin{tabular}{c|c|c|c|c|c}
        \toprule
            Dataset & Animal-Animal & Animal-Obj & Obj-Obj & SSD & TSD \\ \midrule 
            \# of prompts & 66 & 144 & 66 & 31 & 21 \\ \midrule 
            \# of images& 4290 & 9360 & 4290 & 2015 & 1365\\ \bottomrule
        \end{tabular}}
        \vspace{-2mm}
        \caption{Number of prompts and images for each dataset.}
        \vspace{-3mm}
    \label{tab:dataset}
    \end{table}

\section{An alternative loss between aggregated self-attention maps}
\begin{figure}
\centering
\begin{tabular}{ccc}
    \includegraphics[scale=0.26]{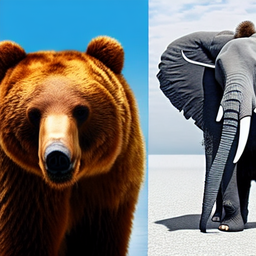} & \includegraphics[scale=0.26]{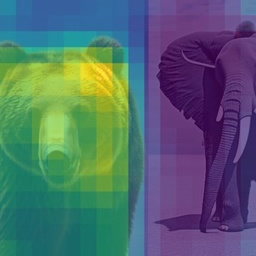} & \includegraphics[scale=0.26]{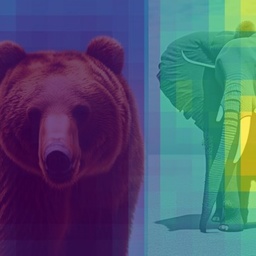} \vspace{-1.5mm} \\
    \footnotesize{image} & \footnotesize{self-attn ``bear''} & \footnotesize{self-attn ``elephant''} 
\end{tabular}
    \vspace{-3mm}
    \caption{\footnotesize{Artiface of image concatenation by Self-Self guidance between the aggregated self-attention maps.}}
    \label{fig:self-self}
    \vspace{-6mm}
\end{figure}

Some readers would suggest an alternative loss to minimize the distance between aggregated self-attention maps(we name it Self-Self guidance in short). Admittedly, this method would achieve comparable results on text-text similarity or TIFA-GPT4o score. However, as shown in Fig.~\ref{fig:self-self}, Self-Self guidance easily leads to the artifact of concatenated images.
While cross-attention maps correspond to subjects only, aggregated self-attention maps can include background.
As Self-Self guidance penalizes any intersection between aggregated self-attention maps, the background is more likely to be separated into two groups resulting in a concatenated image.

\section{Question prompt for TIFA-GPT4o}

In this section, we detail the implementation of TIFA-GPT4o scores and list the full question prompt used in Fig.~\ref{fig:gpt-prompt}. GPT4o’s answers are translated into True (T) or False (F) values for evaluation.

For Existence (Ext), we calculate the percentage of answers when both Question 1 and Question 3 are True. In other words, the presence of both subjects corresponds to the intersection of ``A appears'' and ``B appears''. Similarly, for Recognizability (Rec), we compute the percentage of answers when both Question 2 and Question 4 are True, ensuring that both subjects are recognizable without artifacts or distortions. For Not a Mixture (w/o M), we compute the percentage of answers where Question 5 is False, reflecting the negation of being a mixture.

\begin{figure}
\begin{mdframed}
    You are now an expert to check the faithfulness of the synthesized images. The prompt is \verb|``a {class_A} and a {class_B}''|.
    Based on the image description below, reason and answer the following questions:
\begin{enumerate}
    \item Is there \verb|{class_A}| appearing in this image? Give a True/False answer after reasoning.
    \item Is the generated \verb|{class_A}| recognizable and regular (without artifacts) in terms of its shape and semantic structure only? For example, answer False if a two-leg animal has three or more legs, or a two-eye animal has four eyes, or a two-ear animal has one or three ears. Ignore style, object size in comparison to its surroundings. Give a True/False answer after reasoning.
    \item Is there \verb|{class_B}| appearing in this image? Give a True/False answer after reasoning.
    \item Is the generated \verb|{class_B}| recognizable and regular (without artifacts) in terms of its shape and semantic structure only? For example, answer False if a two-leg animal has three or more legs, or a two-eye animal has four eyes, or a two-ear animal has one or three ears. Ignore style, object size in comparison to its surroundings. Give a True/False answer after reasoning.
    \item Is the generated content a mixture of \verb|{class_A}| and \verb|{class_B}|? An example of mixture is that Sphinx resembles a mixture of a person and a lion. Give a True/False answer after reasoning.
\end{enumerate}
\end{mdframed}
\vspace{-5mm}
\caption{Our Question Prompt for TIFA-GPT4o. Question 1 \& 3 ask about the existence of objects; Question 2 \& 4 ask about the recognizability of objects; Question 5 asks about whether the generated content resembles some mixture of two categories giving the example of Sphinx as in-context learning.}
\label{fig:gpt-prompt}
\end{figure}

\section{Unreliability of CLIP scores}

\begin{figure}[t]
\centering
    \scalebox{0.9}{
    \begin{tabular}{cc}
        \hspace{-3mm} \includegraphics[width=0.35\columnwidth]{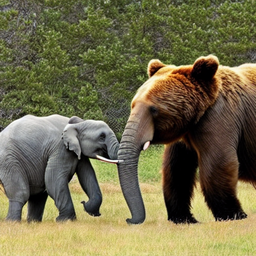} & \hspace{-4mm} \includegraphics[width=0.35\columnwidth]{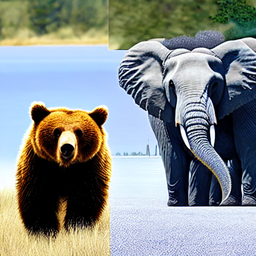} \\ \midrule
        34.4 & 29.8 \\ \midrule
        \hspace{-3mm} \includegraphics[width=0.35\columnwidth]{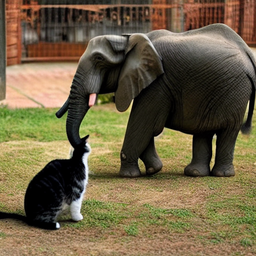} & \hspace{-4mm} \includegraphics[width=0.35\columnwidth]{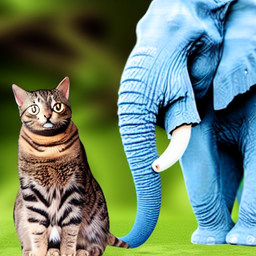} \\ \midrule
        38.0 & 32.6 \\ \midrule
        \hspace{-3mm} \includegraphics[width=0.35\columnwidth]{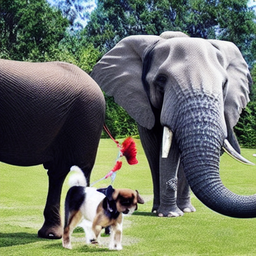} & \hspace{-4mm} \includegraphics[width=0.35\columnwidth]{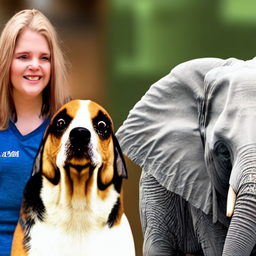} \\ \midrule
        36.0 & 29.6 \\ \bottomrule
    \end{tabular}}
    \caption{CLIP Scores ($\uparrow$) for synthetic images generated by CONFORM~\cite{conform} (left) and our self-cross guidance (right). CLIP scores are unreliable for measuring image quality w.r.t. subject mixing.}
    \label{fig:clip_scores}
\end{figure}

The difference in clip scores between INITNO~\cite{initno}, CONFORM~\cite{conform}, and our method is within 1 \% as shown in Tab.~\ref{tab:min-clip} and~\ref{tab:full-clip}. However, we found CLIP scores unreliable for evaluating the faithfulness of text prompts and synthetic images for subject mixing. Through experiments, we found that the clip score sometimes can't tell subject mixing, as previous work~\cite{tifa} also pointed out. Fig.~\ref{fig:clip_scores} shows example images generated by CONFORM~\cite{conform} and our method with Self-Cross diffusion guidance with the same caption and random seed. For these three pairs of images, Self-Cross diffusion guidance provides visually better images with no subject mixing. However, the corresponding clip scores are much worse than the images generated by CONFORM~\cite{conform}.

\begin{figure}[b] 
    \centering
    \vspace{-3mm}
    \includegraphics[width=0.38\textwidth]{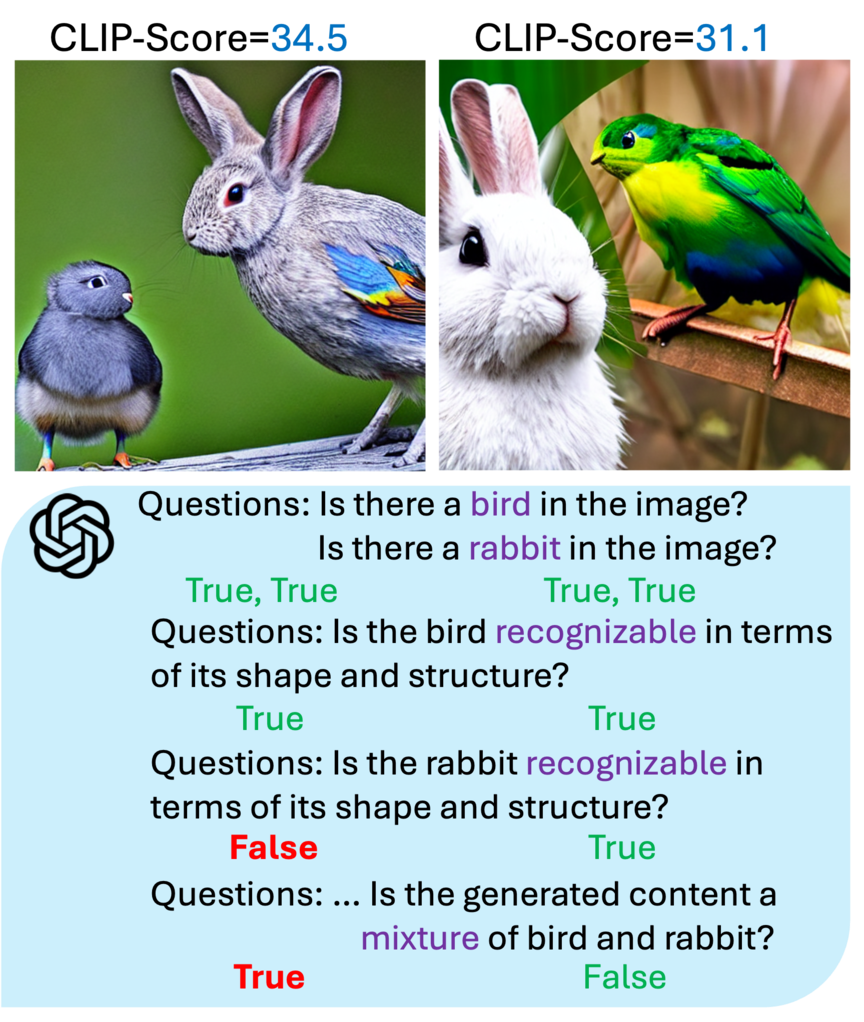} 
    \vspace{-2mm}
    \caption{(Left): image generated by CONFORM \cite{conform}; (right): image generated by our approach under the same seed. Left image shows a higher CLIP score. However, there are obvious content mixing issues in the left image, which GPT4o is able to capture with VQA. This is an example that CLIP score is not as reliable as TIFA for checking subject mixing.}
    \label{fig:clipvstifa}
    \vspace{-6mm}
\end{figure}

Fig.~\ref{fig:clipvstifa} gives a typical example of when the CLIP score is lower for a synthetic image that is more faithful w.r.t. text prompts.
Table.~\ref{tab:reverse-order}.
Tab.~\ref{tab:full-clip} and Tab.~\ref{tab:min-clip} show CLIP scores for different methods with multiple datasets respectively.
While our method outperforms the original stable diffusion for all datasets, it is on par with or slightly worse than other methods in terms of CLIP scores.

    \begin{table}[h]
        \centering
        \footnotesize
        \begin{tabular}{c|c|c|c|c}
        \toprule
            \% & Animal-Animal & Animal-Obj & Obj-Obj & SSD-2 \\ \midrule 
            SD1.4 & 31.0 & 34.3 & 33.6 & 31.2 \\
            INITNO & 33.4 & \textbf{35.9} & \textbf{36.4} & 31.7 \\
            CONFORM & \textbf{33.9} & 35.8 & 35.8 & \textbf{32.0} \\
            Self-Cross & 33.2 & 35.1 & 35.9 & 31.9 \\
            \bottomrule
        \end{tabular}
        \vspace{-3mm}
        \caption{CLIP Scores with full prompts ($\uparrow$) for different methods.}
        \label{tab:full-clip}
        \vspace{-5mm}
    \end{table}
    
    \begin{table}[h]
        \centering
        \footnotesize
        \begin{tabular}{c|c|c|c|c}
        \toprule
            \% & Animal-Animal & Animal-Obj & Obj-Obj & SSD-2 \\ \midrule 
            SD1.4 & 21.6 & 24.8 & 23.9 & 25.8 \\
            INITNO  & 24.9 & \textbf{26.8} & \textbf{27.1} & 26.2 \\
            CONFORM  & \textbf{25.4} & 26.7 & 26.6 & \textbf{26.6} \\
            Self-Cross & 25.1 & 26.1 & 26.7 & \textbf{26.6} \\ \bottomrule
        \end{tabular}
        \vspace{-3mm}
        \caption{CLIP Scores with minimum object prompts ($\uparrow$).}
        \label{tab:min-clip}
        \vspace{-3mm}
    \end{table}
    
    \begin{table}[b]
        \centering
        \resizebox{\columnwidth}{!}{%
        \begin{tabular}{c|c|c|c|c}
        \toprule
            \% & a bear and a turtle & a bird and a bear & a bird and a rabbit & a bird and a lion \\ \midrule 
            original & 35.1 & 33.9 & 32.0 & 33.0  \\
            reverse  & 34.3 & 34.6 & 32.7  & 33.6  \\\bottomrule
        \end{tabular}
        }
        \caption{Inconsistent CLIP Scores $\uparrow$ on a set of images with text prompts reversed.}
        \label{tab:reverse-order}
    \end{table}
    
Additionally, with the same batch of images, the resulting clip score could be different if we simply swap the order of subjects in the prompt during evaluation, as shown in Tab.~\ref{tab:reverse-order}. For example, we generated 65 images with the caption ``a bear and a turtle''. Then we evaluated the clip score with ``a bear and a turtle'' and ``a turtle and a bear'' separately. Surprisingly, we found the clip score for the former is 35.1\% while the clip score for the latter is only 34.3\%. 

To conclude, we resort to the more reliable TIFA-GPT4o scores in this paper, which are more correlated with human judgment, as opposed to the popular CLIP scores.


\section{More qualitative results}

\begin{figure*}
\newcommand{\eachwidth}{0.125\linewidth}
\newcommand{\eachmargin}{\hspace{0mm}}
\newcommand{\pairmargin}{\hspace{-3mm}}
\newcommand{\lefrmargin}{\hspace{-8mm}}
\newcommand{\tltmargin}{\hspace{-4mm}}
\newcommand{\myimgrowOO}[2]{\includegraphics[width=\eachwidth]{sec/images/object-object/sd/#1} & \pairmargin \includegraphics[width=\eachwidth]{sec/images/object-object/sd/#2} & \includegraphics[width=\eachwidth]{sec/images/object-object/initno/#1} & \pairmargin \includegraphics[width=\eachwidth]{sec/images/object-object/initno/#2} & \includegraphics[width=\eachwidth]{sec/images/object-object/conform/#1} & \pairmargin \includegraphics[width=\eachwidth]{sec/images/object-object/conform/#2} & \includegraphics[width=\eachwidth]{sec/images/object-object/selfcross/#1} & \pairmargin \includegraphics[width=\eachwidth]{sec/images/object-object/selfcross/#2}}
\centering
\scalebox{0.87}{
\begin{tabular}{ccc|cc|cc|cc}
& \multicolumn{2}{c|}{SD1.4 \cite{latentdiffusionmodel}} & \multicolumn{2}{c|}{INITNO \cite{initno}} & \multicolumn{2}{c|}{CONFORM \cite{conform}} & \multicolumn{2}{c}{Self-Cross (Ours)}  \\ \midrule \midrule
\lefrmargin \multirow{2}{*}{\rotatebox[origin=c]{90}{\makebox[0pt][c]{\hspace{-1mm}\vspace{2mm}\zapchan{a gray backpack and a green clock}}}} 
& \tltmargin \myimgrowOO{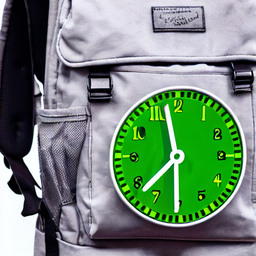}{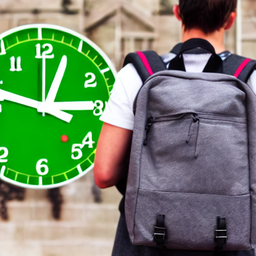} \\
& \tltmargin \myimgrowOO{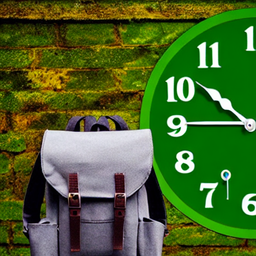}{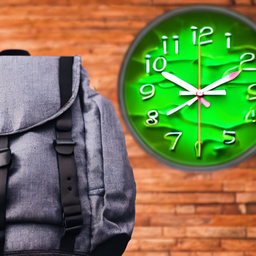} \\ \midrule \midrule
\lefrmargin \multirow{2}{*}{\rotatebox[origin=c]{90}{\makebox[0pt][c]{\hspace{-1mm}\vspace{3mm}\zapchan{a black crown and a red car}}}} 
& \tltmargin \myimgrowOO{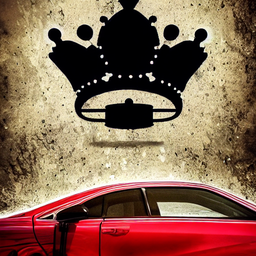}{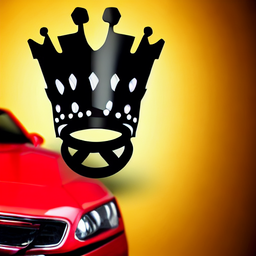} \\
& \tltmargin \myimgrowOO{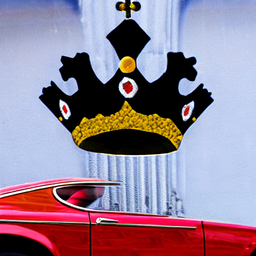}{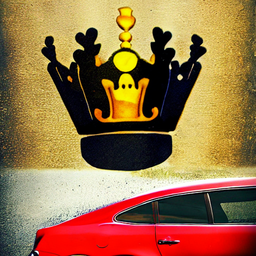} \\ \midrule \midrule
\lefrmargin \multirow{2}{*}{\rotatebox[origin=c]{90}{\makebox[0pt][c]{\hspace{-1mm}\vspace{5mm}\zapchan{a purple chair and a orange bowl}}}} 
& \tltmargin \myimgrowOO{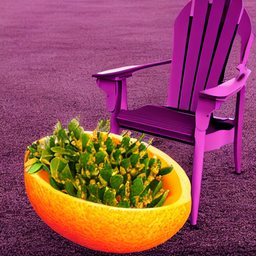}{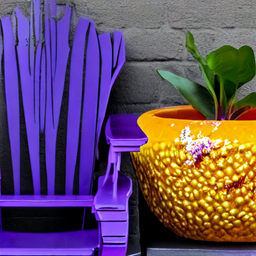} \\
& \tltmargin \myimgrowOO{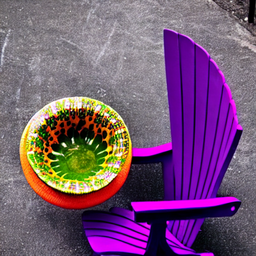}{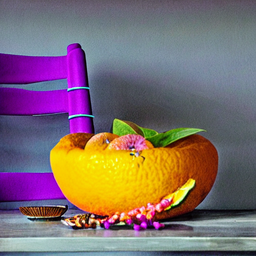} \\ \midrule \midrule
\lefrmargin \multirow{2}{*}{\rotatebox[origin=c]{90}{\makebox[0pt][c]{\hspace{-1mm}\vspace{5mm}\zapchan{a green backpack and a brown suitcase}}}} 
& \tltmargin \myimgrowOO{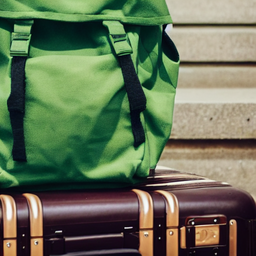}{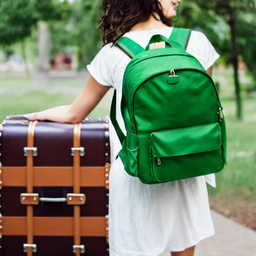} \\
& \tltmargin \myimgrowOO{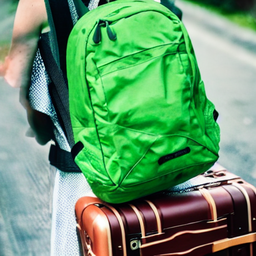}{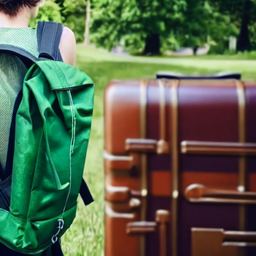} \\ \midrule \midrule
\lefrmargin \multirow{2}{*}{\rotatebox[origin=c]{90}{\makebox[0pt][c]{\hspace{-1mm}\vspace{5mm}\zapchan{a green classes and a black bench}}}} 
& \tltmargin \myimgrowOO{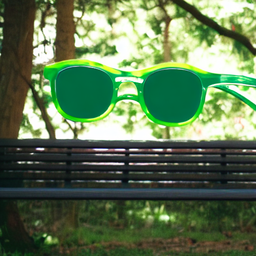}{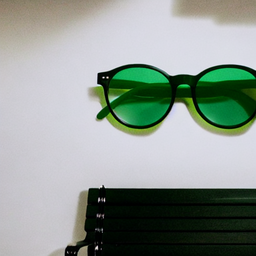} \\
& \tltmargin \myimgrowOO{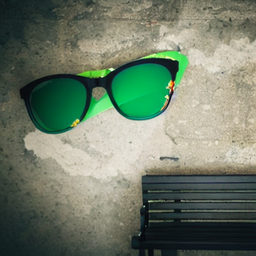}{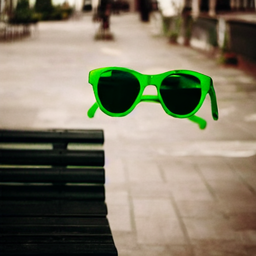}
\end{tabular}}
\vspace{-3mm}
\caption{More qualitative comparisons of Self-Cross (ours) to SD1.4 \cite{latentdiffusionmodel}, INITNO \cite{initno}, CONFORM \cite{conform}. For each prompt in the left column, we sample four seeds and show the results of different methods.}
\label{fig:more-qualitative-comparisons-1}
\vspace{-2mm}
\end{figure*}

\begin{figure*}
\newcommand{\eachwidth}{0.125\linewidth}
\newcommand{\eachmargin}{\hspace{0mm}}
\newcommand{\pairmargin}{\hspace{-3mm}}
\newcommand{\lefrmargin}{\hspace{-8mm}}
\newcommand{\tltmargin}{\hspace{-4mm}}
\newcommand{\myimgrowAA}[2]{\includegraphics[width=\eachwidth]{sec/images/animal-animal/sd/#1} & \pairmargin \includegraphics[width=\eachwidth]{sec/images/animal-animal/sd/#2} & \includegraphics[width=\eachwidth]{sec/images/animal-animal/initno/#1} & \pairmargin \includegraphics[width=\eachwidth]{sec/images/animal-animal/initno/#2} & \includegraphics[width=\eachwidth]{sec/images/animal-animal/conform/#1} & \pairmargin \includegraphics[width=\eachwidth]{sec/images/animal-animal/conform/#2} & \includegraphics[width=\eachwidth]{sec/images/animal-animal/selfcross/#1} & \pairmargin \includegraphics[width=\eachwidth]{sec/images/animal-animal/selfcross/#2}}
\centering
\scalebox{0.87}{
\begin{tabular}{ccc|cc|cc|cc}
& \multicolumn{2}{c|}{SD1.4 \cite{latentdiffusionmodel}} & \multicolumn{2}{c|}{INITNO \cite{initno}} & \multicolumn{2}{c|}{CONFORM \cite{conform}} & \multicolumn{2}{c}{Self-Cross (Ours)}  \\ \midrule \midrule
\lefrmargin \multirow{2}{*}{\rotatebox[origin=c]{90}{\makebox[0pt][c]{\hspace{-1mm}\vspace{2mm}\zapchan{a bear and a lion}}}} 
& \tltmargin \myimgrowAA{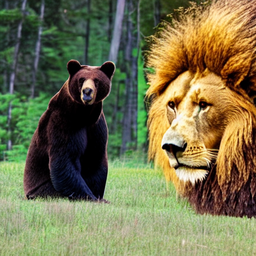}{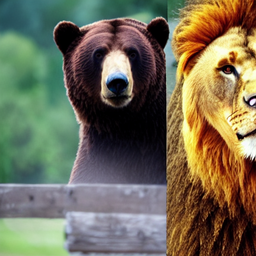} \\
& \tltmargin \myimgrowAA{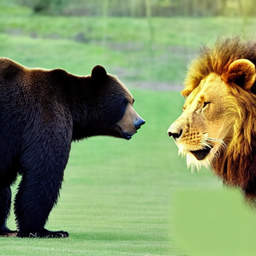}{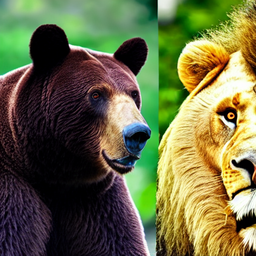} \\ \midrule \midrule
\lefrmargin \multirow{2}{*}{\rotatebox[origin=c]{90}{\makebox[0pt][c]{\hspace{-1mm}\vspace{3mm}\zapchan{a frog and a turtle}}}} 
& \tltmargin \myimgrowAA{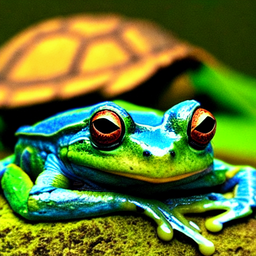}{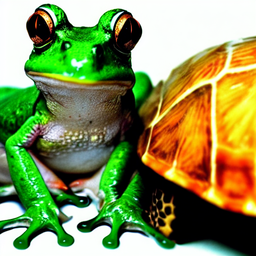} \\
& \tltmargin \myimgrowAA{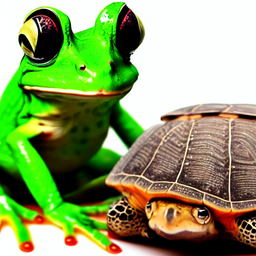}{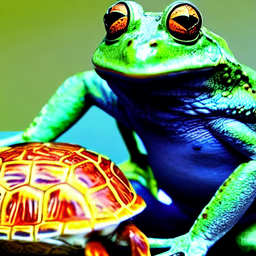} \\ \midrule \midrule
\lefrmargin \multirow{2}{*}{\rotatebox[origin=c]{90}{\makebox[0pt][c]{\hspace{-1mm}\vspace{5mm}\zapchan{a bird and a monkey}}}} 
& \tltmargin \myimgrowAA{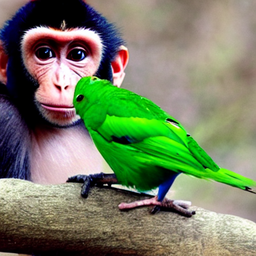}{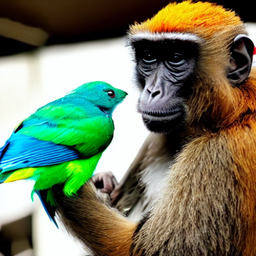} \\
& \tltmargin \myimgrowAA{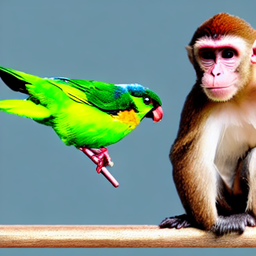}{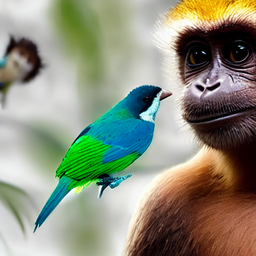} \\ \midrule \midrule
\lefrmargin \multirow{2}{*}{\rotatebox[origin=c]{90}{\makebox[0pt][c]{\hspace{-1mm}\vspace{5mm}\zapchan{a cat and a rabbit}}}} 
& \tltmargin \myimgrowAA{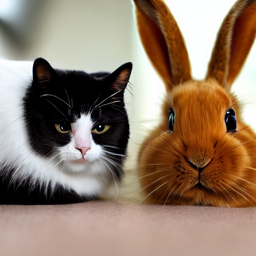}{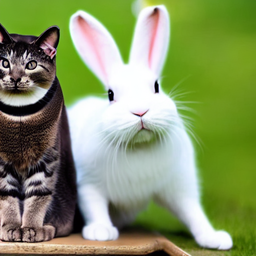} \\
& \tltmargin \myimgrowAA{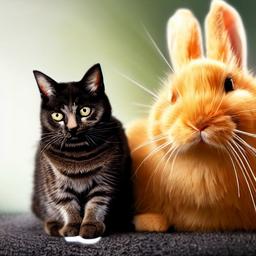}{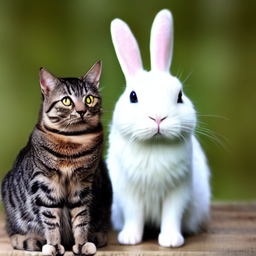} \\ \midrule \midrule
\lefrmargin \multirow{2}{*}{\rotatebox[origin=c]{90}{\makebox[0pt][c]{\hspace{-1mm}\vspace{5mm}\zapchan{a horse and a rat}}}} 
& \tltmargin \myimgrowAA{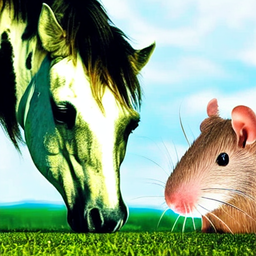}{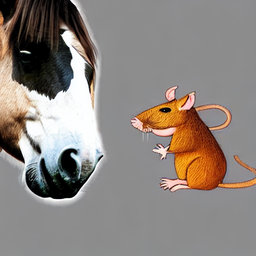} \\
& \tltmargin \myimgrowAA{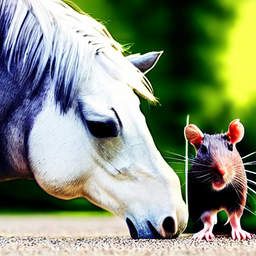}{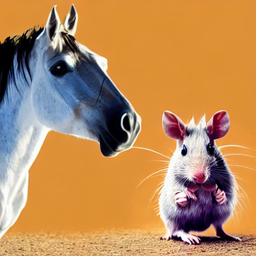}
\end{tabular}}
\vspace{-3mm}
\caption{More qualitative comparisons of Self-Cross (ours) to SD1.4 \cite{latentdiffusionmodel}, INITNO \cite{initno}, CONFORM \cite{conform}. For each prompt in the left column, we sample four seeds and show the results of different methods.}
\label{fig:more-qualitative-comparisons-2}
\vspace{-2mm}
\end{figure*}

We show more qualitative comparisons in Fig.~\ref{fig:more-qualitative-comparisons-1} and Fig.~\ref{fig:more-qualitative-comparisons-2}. 
We select four seeds for each prompt and each method to generate.

These samples illustrate that our approach effectively encourages objects to appear as specified in the prompt. For instance, given the prompt ``a green backpack and a brown suitcase'' in Fig.~\ref{fig:more-qualitative-comparisons-1}, INITNO \cite{initno} sometimes struggles with attribute binding, and CONFORM \cite{conform} often fails to include the `suitcase'. In contrast, our Self-Cross approach successfully addresses these challenges by generating images where both objects are present and correctly aligned with their described attributes.
Moreover, our method excels at resolving subject mixing. Images synthesized using our approach typically feature well-disentangled characteristics for each instance. For example, with the prompt ``a cat and a rabbit'' in Fig.~\ref{fig:more-qualitative-comparisons-2}, other methods often mix features, such as cat faces with rabbit ears, whereas our Self-Cross method accurately generates distinct and faithful representations of both the cat and the rabbit. Similarly, for the prompt a gray backpack and a green clock, other methods sometimes produce ``a green clock-like backpack'', blending features improperly. In contrast, our method faithfully adheres to the prompt, producing clear and visually coherent representations of both the backpack and the clock.

\section{Comparison with Attention Refocusing~\cite{attnrefocusing}}

We further compare our method to Attention Refocusing \cite{attnrefocusing} which depends on external knowledge and model to generate object layout. As shown in Tab.~\ref{tab:attnrefocusing}, our method demonstrates a significant advantage in Existence (Ext), achieving a $7.56\%$ improvement, and an even more substantial advantage in Recognizability (Rec), with a remarkable $23.34\%$ improvement. These results indicate that our approach more effectively ensures that both subjects appear and are free of artifacts or distortions. Additionally, our method achieves comparable performance in reducing subject mixing (w/o M), demonstrating its robustness in separating distinct features of different subjects within the generated images. Our method also shows an improved text-to-text similarity being $4.5\%$ better, which means our generated images are more faithful to the given prompts.

Unlike Attention Refocusing, which relies on a language model to pre-define the layouts, our method operates independently of external knowledge, making it more versatile and applicable to a wider range of scenarios. The superior results in existence and recognizability highlight our approach's ability to generate faithful and high-quality images without relying on external constraints while maintaining competitive performance in mitigating subject mixing.

\begin{table}[h]
    \centering
    \resizebox{\columnwidth}{!}{%
    \begin{tabular}{c|c|c|c}
    \toprule
       Metric ($\uparrow$) & SD1.4 \cite{latentdiffusionmodel} & Attn-Refocus \cite{attnrefocusing} & Self-Cross (Ours) \\ \midrule
        Ext & 39.51 & 86.99 & \textbf{94.55} \\
        Rec & 29.70 & 64.45 & \textbf{87.79} \\
        w/o M & 72.24 & \textbf{93.80} & 92.94 \\ \midrule 
        CLIP score & 31.0 & \textbf{33.9} & 33.2 \\
        Text sim & 76.5 & 79.8 & \textbf{84.3} \\
    \bottomrule
    \end{tabular}
    }
    \caption{Quantitative comparison to Attention Refocusing \cite{attnrefocusing} on Animal-Animal benchmark in terms of TIFA-GPT4o scores \cite{tifa}, CLIP score, and text-to-text similarity (Txt sim) \cite{attendandexcite}. Attention Refocusing relies on external knowledge by using a language model to pre-define the layout. Our proposed method has a significant advantage for existence (Ext), recognizability (Rec), and text-to-text similarity while reaching a comparable performance on reducing subject mixing (w/o M) and CLIP score.}
    \label{tab:attnrefocusing}
\end{table}

\begin{figure}[thbp]
\centering
    \begin{tabular}{c}
        \includegraphics[width=0.25\columnwidth]{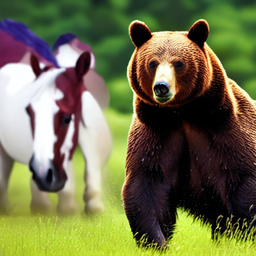}  
        \includegraphics[width=0.25\columnwidth]{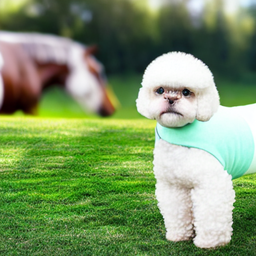} 
        \includegraphics[width=0.25\columnwidth]{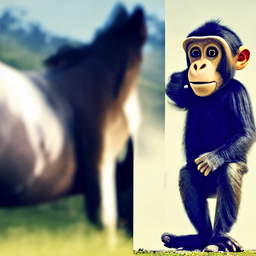} \\ 
        (a) Blurry images \\
        \includegraphics[width=0.25\columnwidth]{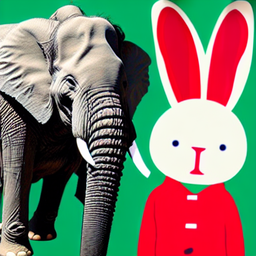}
        \includegraphics[width=0.25\columnwidth]{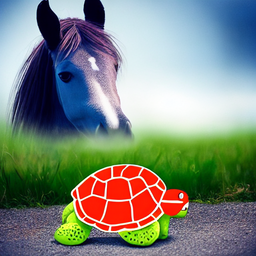}
        \includegraphics[width=0.25\columnwidth]{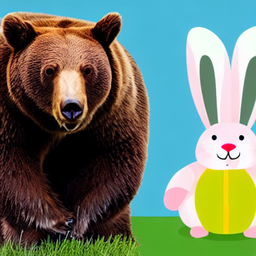} \\ 
        (b) Cartoonish images\\
        \includegraphics[width=0.25\columnwidth]{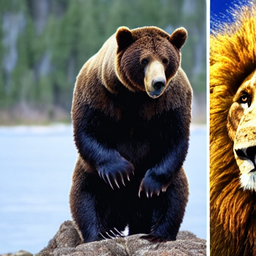}
        \includegraphics[width=0.25\columnwidth]{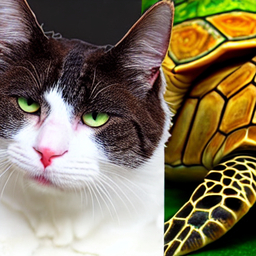}
        \includegraphics[width=0.25\columnwidth]{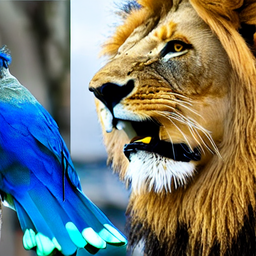} \\ 
        (c) Concatenated subimages.
    \end{tabular}
    \caption{Our method with self-cross guidance failed in some cases and generated blurry images (a), cartoonish images (b), or concatenated subimages (c).}
    \label{fig:failures}
\end{figure}

\section{Failure examples and discussion}
Except for its success in reducing subject mixing, however, Self-Cross Guidance sometimes generates unsatisfactory images, such as blurry images, cartoons, and images with object-centric problems. These failure cases indicate that the method is not perfect. We show failure examples of our method in Fig.~\ref{fig:failures}. We suspect that the artifact of blurriness can be addressed by aggregation of attention maps at higher resolution. We also found that previous methods including INITNO~\cite{initno} and CONFORM~\cite{conform} may also produce cartoonish or concatenated images.

    





\clearpage

